\documentclass[preprint,3p,times,twocolumn,sort&compress]{elsarticle}

\pdfoutput=1

\journal{NEUROCOMPUTING}

\usepackage{fixltx2e}

\usepackage{url}
\usepackage[breaklinks]{hyperref}

\usepackage{algorithmic}
\usepackage{algorithm}

\usepackage{amsmath}
\usepackage{mathtools}

\usepackage{multirow}
\usepackage{booktabs}
\usepackage{makecell}

\usepackage[format=hang, labelformat=parens]{subcaption}
\begin{document}

\begin{frontmatter}

\title{Using Spatial Pooler of Hierarchical Temporal Memory to classify noisy videos with predefined complexity}
\author[aghaddress,cyfronetaddress]{Maciej Wielgosz}
\ead{wielgosz@agh.edu.pl}

\author[aghaddress,cyfronetaddress]{Marcin Pietro\'n}
\ead{pietron@agh.edu.pl}

\address[aghaddress]{AGH University of Science and Technology, Krak\'ow, Poland}
\address[cyfronetaddress]{Academic Computer Centre CYFRONET, Krak\'ow, Poland}

\begin{abstract}
This paper examines the performance of a Spatial Pooler (SP) of a Hierarchical Temporal Memory (HTM) in the task of noisy object recognition. To address this challenge, a dedicated custom-designed system based on the SP, histogram calculation module and SVM classifier was implemented. In addition to implementing  their own version of HTM, the authors also designed a profiler which is capable of tracing all of the key parameters of the system. This was necessary, since an analysis and monitoring of the system performance turned out to be extremely difficult using conventional testing and debugging tools.

The system was initially trained on artificially prepared videos without noise and then tested with a set of noisy video streams. This approach was intended to mimic a real life scenario where an agent or a system trained to deal with ideal objects faces a task of classifying distorted and noisy ones in its regular working conditions.

The authors conducted a series of experiments for various macro parameters of HTM SP, as well as for different levels of video reduction ratios. The experiments allowed them to evaluate the performance of two different system setups (i.e. 'Multiple HTMs' and 'Single HTM') under various noise conditions with 32--frame video files. Results of all the tests were compared to SVM baseline setup.

It was determined that the system featuring SP is capable of achieving approximately 12 times the noise reduction for a video signal with with distorted bits accounting for 13\% of the total. Furthermore, the system featuring SP performed better also in the experiments without a noise component and achieved a max F1 score of 0.96. 
The experiments also revealed that a rise of column and synapse number of SP has a substantial impact on the performance of the system. Consequently, the highest F1 score values were obtained for 256 and 4096 synapses and columns respectively. It is worth noting that 'Multiple HTMs' setup outperforms the 'Single HTM' one. Nevertheless, this is achieved at the expense of increased computational demands. 
 
\end{abstract}


\end{frontmatter}


\section{Introduction}
\label{section:intro}

\nocite{wielgosz2016using}

The world consists of objects that interact dynamically with each other to form a network of complex relationships. It should be noted that virtually all processes in the real world develop in time; even the ones that appear to be fixed, after careful analysis in the micro--scale, turn out to be dynamic.

The eyes are the basic and primary tool for examining the world with which the nature equipped virtually all the living creatures capable of moving by themselves. 
Closer examination of species' evolution on Earth show that a huge leap forward occurred around the time when the fist primitive eyes were developed which happened about 540 million years ago \cite{wiki:Eye}.  

Stationary images, even ones of a very high resolution and great number of low level details, provide limited information regarding an analyzed scene and objects within it \cite{Han2015Object, Cheng2015Effective, Du2014Discriminative, Zhang2015Saliency}. They lack a temporal component that is essential to follow how the objects change in time, which in turn provides complete information of its nature. Furthermore, it is worth noting that the human brain was well adapted for processing dynamic object interactions. The important role that the dimension of time plays in analyzing visual objects is also reflected in the tissue structure, 80\% of which involves temporal connections \cite{Mountcastle}. Consequently, when analyzing static scenes our eyes make rapid, step--like rotation, following which the eyes remain stationary. The series of step movements are known as saccades or saccadic eye movements. 

The real world is highly noisy and imperfect, thus the brain was designed to analyze such distorted streams of images. It is able to learn to distinguish between heavily malformed and noisy objects in the space we live in. This is possible due to a long and complicated process of training, during which the brain is exposed to streams of data in large quantities and learns to extract the perfect shapes of objects from the noisy ones. The process may be perceived as discovering latent features which are unique and invariant for a given object or set of objects. The idea behind this approach resembles to some extent the methods used for training the latest deep learning algorithms. A sufficiently large amount of stationary images are used to compensate for the lack of a temporal component in exploring latent features of the processed objects. This phenomenon contributed to the recent breakthrough in Very Deep Learning architecture training which is mostly driven by the availability of a huge amount of data and GPU computational power \cite{Krizhevsky, Schmidhuber}.

The brain, in an effort to classify distorted images, compares the ideal version of the objects encoded in Sparse Distributed Representation (SDR) with the noisy ones. The authors decided to investigate to what extent this mechanism is effectively implemented in Hierarchical Temporal Memory (HTM) \cite{wielgosz2016using}, the bio-inspired algorithm, which models a selected section of neocortex \cite{Numenta} using state--of--the--art knowledge in neurobiology regarding the human brain.

The hypothesis which the paper attempts to confirm states that \textit{incorporating SP of HTM in video processing flow increases the overall noise robustness of system}.

The human brain as a whole has not been completely explored yet, making its artificial implementation and verification a very hard task. However, there are initiatives \cite{humanbrainproject} which have taken up the challenge of simulating and modeling the brain as we know it today. Rather than model the brain, the authors of this paper have adopted a slightly different approach of gradually introducing selected components of Hierarchical Temporal Memory (HTM) to the video processing system with the intention of enhancing its performance. By doing so we aim to develop a complete system\cite{online:custom_htm} working on the principles of the human brain as they were presented in \cite{Mountcastle, Numenta} with our modification making the algorithm suitable for hardware implementation. Running HTM on CPU is very slow, and due to its strongly parallel structure the algorithm is a good candidate for General--Purpose Graphics Processing Unit (GPGPU) and Field--Programmable Gate Array (FPGA) acceleration \cite{pietron2016parallel, pietron2016formal}. Therefore, the computationally demanding overlap and inhibition sections of SP were implemented on GPU. 

The paper contains the following four main contributions:
\begin{itemize}
 \item analysis of noise reduction properties of SP in video processing using object detection as a case study setup,
 \item experimental confirmation of SP noise reduction properties,
 \item experimental study of the impact of key parameter changes
 on the performance of the SP-based system,
 \item development of a custom-designed prototype application of an HTM solution meant for doing video experiments, along with a data set. The software and the video data is available online at \cite{online:custom_htm, online:datasets}.
\end{itemize}

The rest of the paper is organized as follows. Sections \ref{subsection:object_detection} and \ref{subsection:HTM} provide the background and related work of  object classification in video streams and Hierarchical Temporal Memory, respectively. Section \ref{section:sp_processing} contains mathematical analysis of the noise reduction properties of SP. Architecture of the custom--designed system used for the experiments is described in Section \ref{section:system_description}, with the data flow presented in Section \ref{section:processing_flow}. Section \ref{section:experiments} provides the results of the experiments. Finally, the conclusions of our research are presented in Section \ref{section:conclusions}.

\subsection{Object classification in video streams}
\label{subsection:object_detection}

\begin{figure}
\centering
\includegraphics[width=0.45\textwidth]{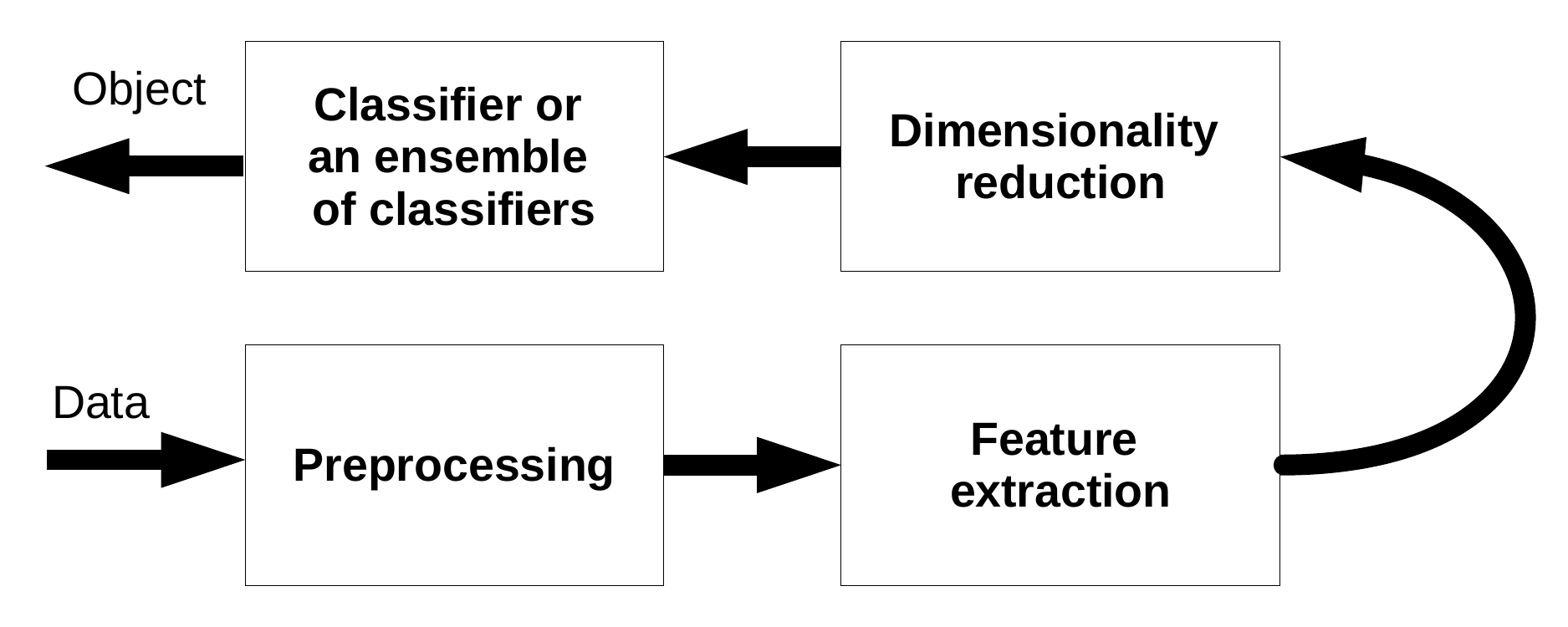}
\caption{Architecture of a video processing system}
\label{fig:architecture_of_a_video_processing_system}
\end{figure}

Most state--of--the--art information extraction systems consist of the following sections: preprocessing, feature extraction, dimensionality reduction and classifier or ensemble of classifiers (Fig. \ref{fig:architecture_of_a_video_processing_system}). Their construction requires expert knowledge as well as familiarity with the data that will be processed \cite{Haibo, Peng}. 

Usually, systems for object classification in video streams are also designed according to this scheme. Consequently, the proper choice of the operations which constitute all the mentioned stages of the system is important and determines the classification result \cite{Lu, Hota, Islam}. One of the most challenging stages is feature extraction, which substantially affects the overall performance of the system.  

There are also systems which take advantage of the spatial--temporal \cite{Numenta} profile  of the data\cite{Castrill,Devarakota,Khan,Bengio}. They are closer to the concept of the solution presented in this paper, which may be considered a hybrid approach since it features components of both schemes.

\subsection{Hierarchical Temporal Memory}
\label{subsection:HTM}

Hierarchical Temporal Memory (HTM) replicates the structural and algorithmic properties of the neocortex. It can be regarded as a memory system which is not programmed, but trained through exposing it to data flow. The process of training is similar to the way humans learn which, in its essence, is about finding latent causes in the acquired content. At the beginning, the HTM has no knowledge of the data stream causes it examines, but through a learning process it explores the causes and captures them in its structure. The training is considered complete when all the latent causes of data are captured and stable. The detailed presentation of HTM is provided in \cite{Numenta, Chen, Rachkovskij}.

HTM constitutes a hierarchy of nodes, where each node performs the same algorithm. The most basic elements (raw and unprocessed data) enter at the bottom of the hierarchy. Each node learns the spatio--temporal pattern of its input and associates it with a given concept. Consequently, each node, no matter where it is in the hierarchy, discovers the causes of its input.
In an HTM, beliefs exist at all levels in the hierarchy and are internal states of each node. They represent probabilities that a cause is active. Each node in an HTM has a fixed number of concepts and a fixed number of output variables. The training process of an HTM starts with a fixed number of possible causes, and in a training process, assigns a meaning to them.

Consequently, the nodes do not increase the number of concepts they cover; instead, over the course of the training, the meaning of the outputs gradually changes. This happens at all levels in the hierarchy simultaneously. Thus the top level of the hierarchy remains with little or no meaning till nodes at the bottom are trained to recognize the basic patterns.

HTM is composed of two main parts, namely Spatial and Temporal Pooler (TP). This paper focuses on Spatial Pooler (SP), aka Pattern Memory, which is employed in the processing flow of the system. It contains columns with synapses connected to the input data \cite{Numenta}. The main role of SP in HTM is finding spatial patterns in the input data. It may be decomposed into three stages:

\begin{itemize}
 \item Overlap calculation (Alg. \ref{alg:overlap}),
 \item Inhibition (Alg. \ref{alg:inhibition}),
 \item Learning.
\end{itemize}

\begin{algorithm}[t]
\caption{Overlap }
\label{alg:overlap}
\begin{algorithmic}[1]
\FORALL{col $\in$ sp.columns} 
\STATE{col.overlap $\leftarrow$ 0} 

\FORALL{syn $\in$ col.connected\_synapses()} 
\STATE{col.overlap $\leftarrow$ col.overlap + syn.active()} 
\ENDFOR

\IF{col.overlap $<$ min\_overlap}
\STATE col.overlap $\leftarrow$ 0
\ELSE
\STATE col.overlap $\leftarrow$ col.overlap * col.boost
\ENDIF
\ENDFOR
\end{algorithmic}
\end{algorithm}

\begin{algorithm}[tbp]
\caption{Inhibition}
\label{alg:inhibition}
\begin{algorithmic}[2]
\FORALL{col $\in$ sp.columns} 
\STATE{max\_column $\leftarrow$ max(n\_max\_overlap(col, n), 1)} 
\STATE col.active $\leftarrow$ col.overlap $>$ max\_column
\ENDFOR
\end{algorithmic}
\end{algorithm}

The first two stages are very computationally demanding but can be parallelized. Therefore the authors decided to implement them on GPU in OpenCL \cite{wielgosz2016opencl}. The learning stage, the detailed description of which is provided in the Numenta whitepaper \cite{Numenta}, is implemented on CPU.

The overlap section (Alg. \ref{alg:overlap}) computes $col.overlap$ for every column in SP structure i.e. a number of active and connected synapses. If the number is larger than $col.min\_overlap$, then it is boosted and passed on to the inhibition section (Alg. \ref{alg:inhibition}).

The inhibition stage (Alg. \ref{alg:inhibition}) implements a winner--takes--all procedure where for each column a decision is made as to whether it belongs to a range of $n$ ($winners\_set\_size$) columns of the highest values. The $n\_max\_overlap()$ function performs the comparison.

\section{SP processing}
\label{section:sp_processing}

\begin{figure*}
\centering
    \includegraphics[width=0.94\textwidth]{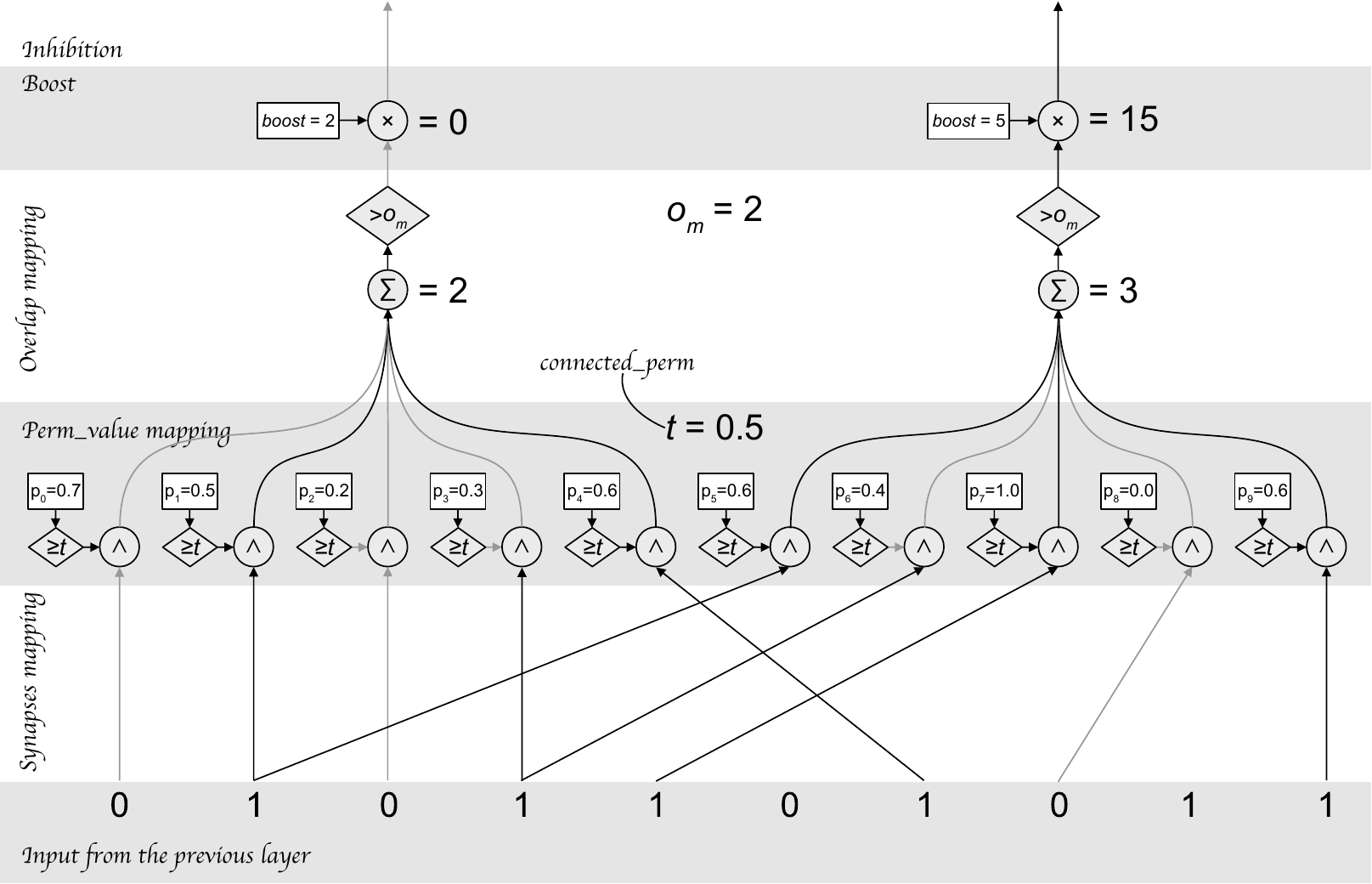}
    \caption{Spatial Pooler architecture}
    \label{fig:sp_architecture}
\end{figure*}

This section covers properties and important features of the SDR (Sparse Distributed Representation) vector space as well as presents SP processing flow as a set of steps. Those steps may be described as mappings between spaces which are progressively narrowed to extract features to be used in the further processing stages that come after SP. 
It is important to note that all the operations presented in this section consider SP after training is accomplished, which means that learning is not taken into account.

\subsection{Notation}

\begin{table}
\caption{Notation}\label{tab:notation}
\centering
\begin{tabular}{cl}
\toprule
$S$ & SDR (Sparse Distributed Representation) \\ \midrule
$d_S(x, y)$ & distance between $x$ and $y$ in SDR \\ \midrule
$n$ & input vector size \\ \midrule
$n_b$ & number of '1' in the input vector \\ \midrule
$w$ & \makecell[l]{number of noise bits\\included in input vector} \\ \midrule
$w_b$ & number of '1' in the noise \\ \midrule
$m$ & $perm\_value$ reduction ratio \\ \midrule
$c$ & number of columns \\ \midrule
$s$ & number of synapses \\ \midrule
$o_m$ & $min\_overlap$ \\ \midrule
$p_x$ & \makecell[l]{$perm\_value$ for synapse\\associated with column input $x$} \\
\bottomrule
\end{tabular}
\end{table}

We use a single notation scheme for math formalism in this section, shown in Tab.~\ref{tab:notation}. The size of all the vectors is expressed in bits. The notation covers all the parameters which are essential for a complete description of the SP architecture. 

Input vector size $n$ along with number of positive bits $n_b$ are considered to be a key parameters since they they define the sparsity of the SP input vector \cite{Ahmad, Numenta}. Number of columns $c$ and number of synapses $s$ directly affects the shape of the SP architecture and decide about its processing capabilities \cite{Cui} such as a noise elimination. The noise content is quantitatively described by $w$ and $w_b$ which shape the statistical distribution of information content carried by the input vectors. It is worth noting that any two vectors may be compared in terms of Hamming measure by applying the distance operator $d_S(x, y)$.

In addition to the basic hyper-parameters of the SP architecture and input vector defining restrictions, Tab.~\ref{tab:notation} also contains more subtle, but also important constants such as $o_m$ and $m$. These constants decide about processing flow of the SP architecture. Min\_overlap sets a threshold for an input data pattern to be regarded as a match. Consequently, $o_m$ affects selectivity of the network. On the other hand $m$ may be perceived as sensitivity shaping parameter.

\subsection{Space definition}
All the operations in the SP processing chain are performed in sparse distributed space.

\begin{equation}
S: \{0,1\}^{1 \times n}
\label{eq:sdr_space}
\end{equation}
where $S$ and $n$ are the SDR vector space and its dimension, respectively.

In video processing the vectors in SDR (Eq. \ref{eq:sdr_space}) may be interpreted as points in $n$ dimensional space. The points are brought to SDR space by an encoding operation which, in this case, is binarization. 
 
\begin{equation}
d_S:S \times S \rightarrow \mathbb{N}
\label{eq:metric_space_distance_metric_S}
\end{equation}

\begin{equation}
\bigwedge_{x, y \in S} d_S(x, y)  \geqslant 0
\label{eq:metric_space_postulate_separation_S}
\end{equation}

\begin{equation}
\bigwedge_{x, y \in S} d_S(x, y) = 0 \Longleftrightarrow x = y
\label{eq:metric_space_identity_S}
\end{equation}

\begin{equation}
\bigwedge_{x, y \in S} d_S(x, y) = d_S(y, x)
\label{eq:metric_space_symmetry_S}
\end{equation}

\begin{equation}
\bigwedge_{x, y, z  \in S} d_S(x, z)  \leqslant d_S(x, y) + d_S(y, z)
\label{eq:metric_space_triangle_S}
\end{equation}

\subsection{SP mapping}
The Spatial Pooler processing chain may be decomposed as a series of mapping operations. This allows for analysis of data structure and noise contribution at each stage separately. Regardless of the size of the input data and noise volume, the operations have the same structure.

\subsubsection{Synapse connectivity mapping}
In the first step every column maps from the input space $S_0$ to its own subspace $S_1$. The operation is described by Eq. \ref{eq:synapse_connectivity_mapping}.

\begin{equation}
S_0: \{0,1\}^{1 \times n} \xRightarrow{\Pi_0} S_1: \{0,1\}^{s \times c}
\label{eq:synapse_connectivity_mapping}
\end{equation}

There is a fixed number of choices every column makes, which is determined by its synapse number as expressed by Eq. \ref{eq:synapse_connectivity_mapping_choices}.

\begin{equation}
\binom {n}{s}=\frac{n!}{s!(n-s)!}
\label{eq:synapse_connectivity_mapping_choices}
\end{equation}

\subsubsection{Perm\_value mapping}
Once a subset of input data is selected by each column, the $perm\_value$ mapping operation brings a vector from synapse to $perm\_value$ space i.e. from $S_1$ to $S_2$ as given by Eq.~\ref{eq:synapse_perm_mapping}. This may reduce to some extent the size of the data vector to be processed.

\begin{equation}
S_1: \{0,1\}^{s \times c} \xRightarrow{\Pi_1} S_2: \{0,1\}^{s \times c}
\label{eq:synapse_perm_mapping}
\end{equation}

$Perm\_values$ associated with every synapse of each column ($p_x$) are compared against $connected\_perm$. When an element of $S_1$ interacts with a connected synapse i.e. for which $perm\_value \geqslant connected\_perm$, it is included in the $S_2$ space as expressed by Eq. \ref{eq:synapse_perm_inclusion_condition}.

\begin{equation}
\Pi_1: \bigwedge_{x \in S_1} x \in S_2 \Longleftrightarrow x = 1 \land p_x \geqslant connected\_perm
\label{eq:synapse_perm_inclusion_condition}
\end{equation}

\subsubsection{Overlap mapping}
In this step all the bits which passed through $\Pi_0$ and $\Pi_1$ mapping are counted to obtain an overlap value which may be again viewed as the mapping operation that is given by Eq. \ref{eq:overlap_mapping}.

\begin{equation}
S_2: \{0,1\}^{s \times c} \xRightarrow{\Pi_2} S_3: \{0, \ldots , s\}^{1 \times c}
\label{eq:overlap_mapping}
\end{equation}

\begin{equation}
\Pi_2 : 
\bigwedge_{\substack{i \in \left\langle 0, s \right) \\ \land \\ j \in \left\langle 0, c \right)}} \bigvee_{\substack{x_{i,j} \in S_2 \\ \land \\ y_j \in S_3}} y_j = \sum_{i =0}^{s} x_{i,j}
\label{eq:overlap_condition}
\end{equation}

Once the overlap value is calculated according to Eq.~\ref{eq:overlap_condition}, it is compared against $min\_overlap$ as given by Eq.~\ref{eq:min_overlap_condition}. Based on the comparison, the column is either included in the active columns set or remains excluded from further processing.

\begin{equation}
S_3: \{0, \ldots, s\}^{1 \times c} \xRightarrow{\Pi_3} S_4: \{0,1\}^{1 \times c}
\label{eq:min_overlap_mapping}
\end{equation}

\begin{equation}
\Pi_3 : 
\bigwedge_{x \in S_3} \bigvee_{y \in S_4} y =
\begin{cases} 
1 \text{~if~} x > o_m \\
0 \text{~if~} x \leqslant o_m \end{cases}
\label{eq:min_overlap_condition}
\end{equation}

\subsubsection{Boost}

Conducted experiments show that boost barely affects the activity pattern of SP columns. Therefore, authors decided not to include it in the formal analysis. This makes the formalism simpler and easier to follow and also directs the focus to the most contributive sections of the algorithm.

\subsubsection{Inhibition}
Each column is selected as either active or inactive depending on its overlap value in the context of all the neighboring columns. The comparison range is determined by the inhibition range \cite{Numenta, How_neurons}.

According to authors' experimental observations (Fig. \ref{fig:visualizer_inhibition_radius}) the inhibition radius does not change after the SP has been trained. Consequently, a small amount of noise in the input will change the output to some extent, but most of the winners will remain unchanged.

\subsection{Signal propagation}
The choice of the winning column set depends ultimately on the number of '1' in the input vector ($n_b$). Therefore we decided to analyze the impact of this number and SP parameters on the selectivity of SP.

\subsubsection{Synapse connectivity mapping}
The number of '1' contained within a random vector in $S_1$ subspace may be expressed in terms of the random variable $X_{\text{SCM}}$ with binomial distribution given by Eq.\ref{eq:binomial_signal_synapse_mapping}.

\begin{equation}
E[X_{\text{SCM}}] = \sum_{k=1}^{s}k \binom {s}{k}{\left( \frac{n_b}{n} \right)}^k \left(1-\frac{n_b}{n}\right)^{(s-k)}= s\frac{n_b}{n}
\label{eq:binomial_signal_synapse_mapping}
\end{equation}

where $E[X_{\text{SCM}}]$ is the expected value of the random variable.

\subsubsection{Perm\_value mapping}
The number of '1' contained within a random vector in the subspace $S_2$ may be expressed in terms of the random variable $X_{\text{SPM}}$ with binomial distribution given by Eq. \ref{eq:binomial_signal_perm_value_mapping}.

\begin{equation}
\begin{split}
E[X_{\text{SPM}}] & = \sum_{k=1}^{s m}k \binom {s m}{k}{\left(\frac{E[X_{\text{SCM}}]}{s}\right)}^k \left(1-\frac{E[X_{\text{SCM}}]}{s}\right)^{(s m - k)} \\
                            & = s m \frac{E[X_{\text{SCM}}]}{s} = s m \frac{n_b}{n}
\end{split}
\label{eq:binomial_signal_perm_value_mapping}  
\end{equation}

where $E[X_{\text{SPM}}] $ is the expected value of the random variable. The $m$ parameter ($perm\_value$ reduction ratio) can be expressed as in Eq. \ref{eq:m_definition}.

\begin{equation}
\bigwedge_{x \in S_1 ~\land ~y \in S_2} m = \frac{|y|}{|x|}
\label{eq:m_definition}
\end{equation}

\subsubsection{Overlap mapping}
The main goal of using SP in data processing systems regardless of the target application is improving feature extraction quality through increasing pattern matching accuracy. This ability of filtering out the right candidates for column activation is crucial. It may be described in terms of the probability of matching between a random input vector and synapse connection pattern in $S_2$ space.

\begin{equation}
P_{n_b}^n(o_m)_{\text{signal}} = \frac {\sum_{k=o_m}^s\binom {n_b}{k} \times \binom{n-n_b}{E[X_{\text{SPM}}]}}{\binom{n}{E[X_{\text{SPM}}]}}
\label{eq:overlap_signal_mapping} 
\end{equation}

The detailed description on how Eq. \ref{eq:overlap_signal_mapping} was derived is provided in \cite{How_neurons}.

An increase of $n_b$, $n$ and $o_m$ leads more to the exponential growth of the denominator rather than the numerator. This, in turn, results in a drop of the false positives ratio. 

\subsection{Noise propagation}
Noise introduced to the input vector is propagated through all the mapping steps contributing to signal distortion.

\subsubsection{Synapse connectivity mapping}
The number of noise bits contained within in the subspace $S_1$ may be expressed in terms of the random variable $X_{\text{NCM}}$ with binomial distribution given by Eq.~\ref{eq:binomial_noise_synapse_mapping}.

\begin{equation}
E[X_{\text{NCM}}] = \sum_{k=1}^{s}k \binom {s}{k}{\left(\frac{w}{n}\right)}^k \left(1-\frac{w}{n}\right)^{(s-k)}= s\frac{w}{n}
\label{eq:binomial_noise_synapse_mapping}
\end{equation}

where $E[X_{\text{NCM}}] $ is the expected value of the random variable.

\subsubsection{Perm\_value mapping}
The number of noise bits contained within a random vector in the subspace $S_2$ may be expressed in terms of the random variable $X_{\text{NPM}}$ with binomial distribution given by Eq.~\ref{eq:binomial_noise_perm_value_mapping}.

\begin{equation}
\begin{split}
E[X_{\text{NPM}}]  & = \sum_{k=1}^{s m}k \binom {s m}{k}{\left(\frac{E[X_{\text{NCM}}]}{s}\right)}^k \left(1-\frac{E[X_{\text{NCM}}]}{s}\right)^{(s m-k)} \\
                             & = s m \frac{E[X_{\text{NCM}}]}{s} = s m \frac{w}{n}
\end{split}
\label{eq:binomial_noise_perm_value_mapping} 
\end{equation}

where $X_{\text{NPM}}$ is the random variable that takes $w$ number of $k$ noise bits for each column after $perm\_value$ mapping.

\subsubsection{Overlap mapping}
Noise introduction affects the number of '1' in the data vector mapped to $S_2$ which may be expressed in terms of the expected value of a random variable with binomial distribution given by Eq.\ref{eq:binomial_noise_overlap_mapping}.

\begin{equation}
\begin{split}
E[X_{\text{NB}}] & = \frac {w_b}{n} s + \left(1 - \frac{w}{n}\right)\frac{n_b}{n} s \\
                          & = \frac{s}{n}\left[w_b + n_b \left(1 - \frac{w}{n}\right)\right]
\end{split}
\label{eq:binomial_noise_overlap_mapping} 
\end{equation}

The first part of Eq. \ref{eq:binomial_noise_overlap_mapping}: $\frac {w_b}{n} s$  accounts for noise contribution, whereas the second part : $\left(1 - \frac{w}{n}\right)\frac{n_b}{n} s$ covers all the '1' included in the remaining part of the signal which was not affected by the noise.

\begin{equation}
P_{n_b}^n(o_m)_{\text{noise}} = \frac {\sum_{k=o_m}^s\binom {n_b}{k} \times \binom{n-n_b}{E[X_{\text{NB}}]}}{\binom{n}{E[X_{\text{NB}}]}}
\label{eq:overlap_noise_mapping} 
\end{equation}

\subsection{Noise impact}
The noise impact on the false positive ratio is given by the Eq. \ref{eq:noise_impact}.

\begin{equation}
\begin{split}
\frac{P_{n_b}^n(o_m)_{\text{signal}}}{P_{n_b}^n(o_m)_{\text{noise}}}  = \frac {\sum_{k=o_m}^s\binom {n_b}{k} \binom{n-n_b}{E[X_{\text{SPM}}]}\binom{n}{E[X_{\text{NB}}]}}{\sum_{k=o_m}^s\binom {n_b}{k} \binom{n-n_b}{E[X_{\text{NB}}]} \binom{n}{E[X_{\text{SPM}}]}} \\
= \frac{(n - n_b - E[X_{\text{NB}}])! (n - E[X_{\text{SPM}}])!}{(n - n_b - E[X_{\text{SPM}}])! (n - E[X_{\text{NB}}])!} \sum_{k=o_m}^s\binom {n_b}{k}
\end{split}
\label{eq:noise_impact} 
\end{equation}

The equation presents the degree of ambiguity of pattern matching results caused by a detrimental noise contribution.

It is worth noting that false negative cases are also possible but this effect may be neglected with relatively low $min\_overlap$ values. The detailed study of this issue is available in \cite{How_neurons}. 
\section{System description}
\label{section:system_description}
The custom-designed system \cite{online:custom_htm} is highly configurable, with numerous parameters responsible for the core HTM's structure, the encoder behavior, statistics rendering, etc. In addition to the core module, a set of supporting modules has been developed. Most of them are used for feeding video data to the core module, and receiving and analyzing the results. 

\begin{figure*}
\centering
\begin{subfigure}{0.48\textwidth}
    \centering
    \includegraphics[scale=0.6]{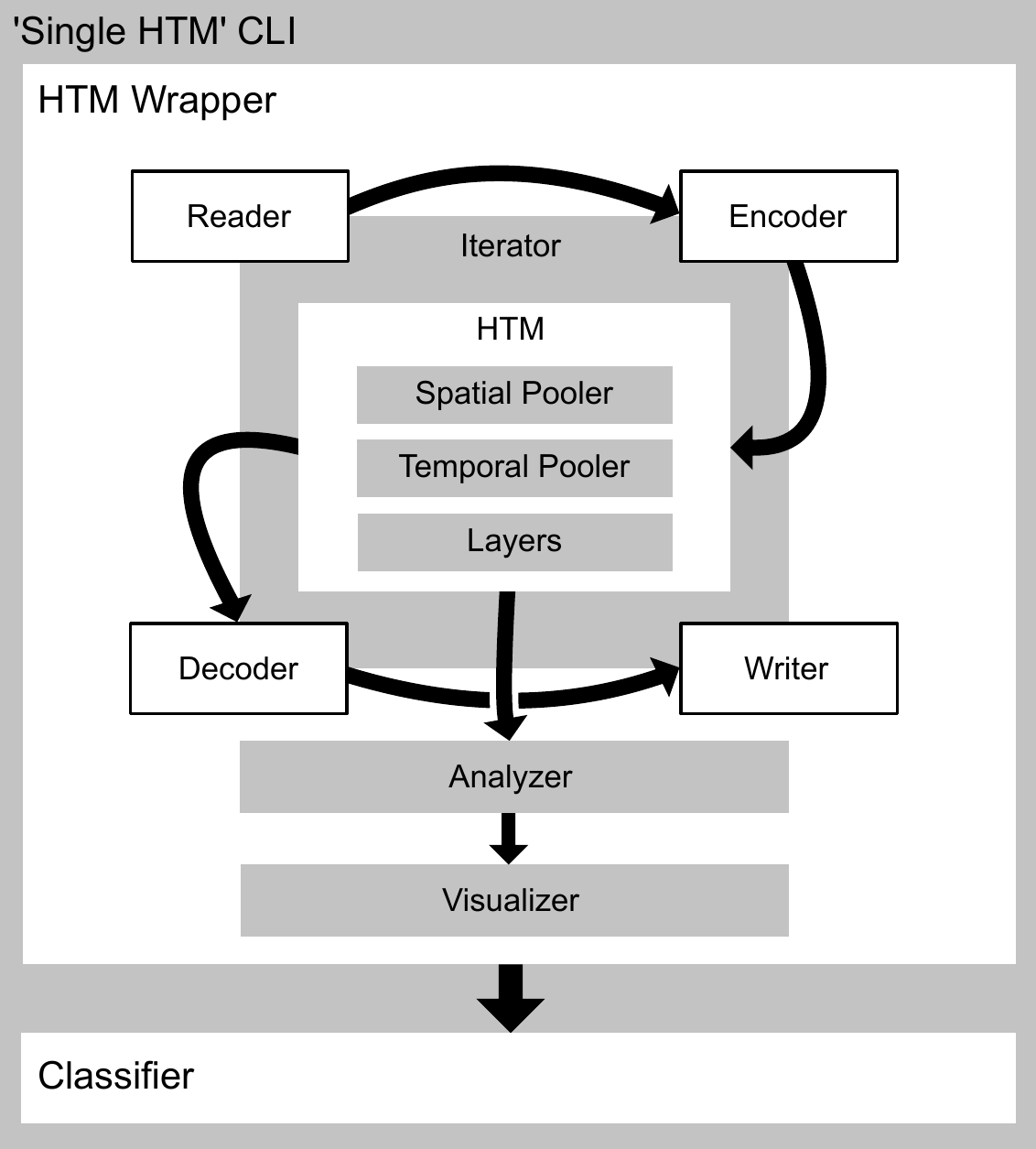}
\end{subfigure}
\begin{subfigure}{0.48\textwidth}
    \centering
    \includegraphics[scale=0.6]{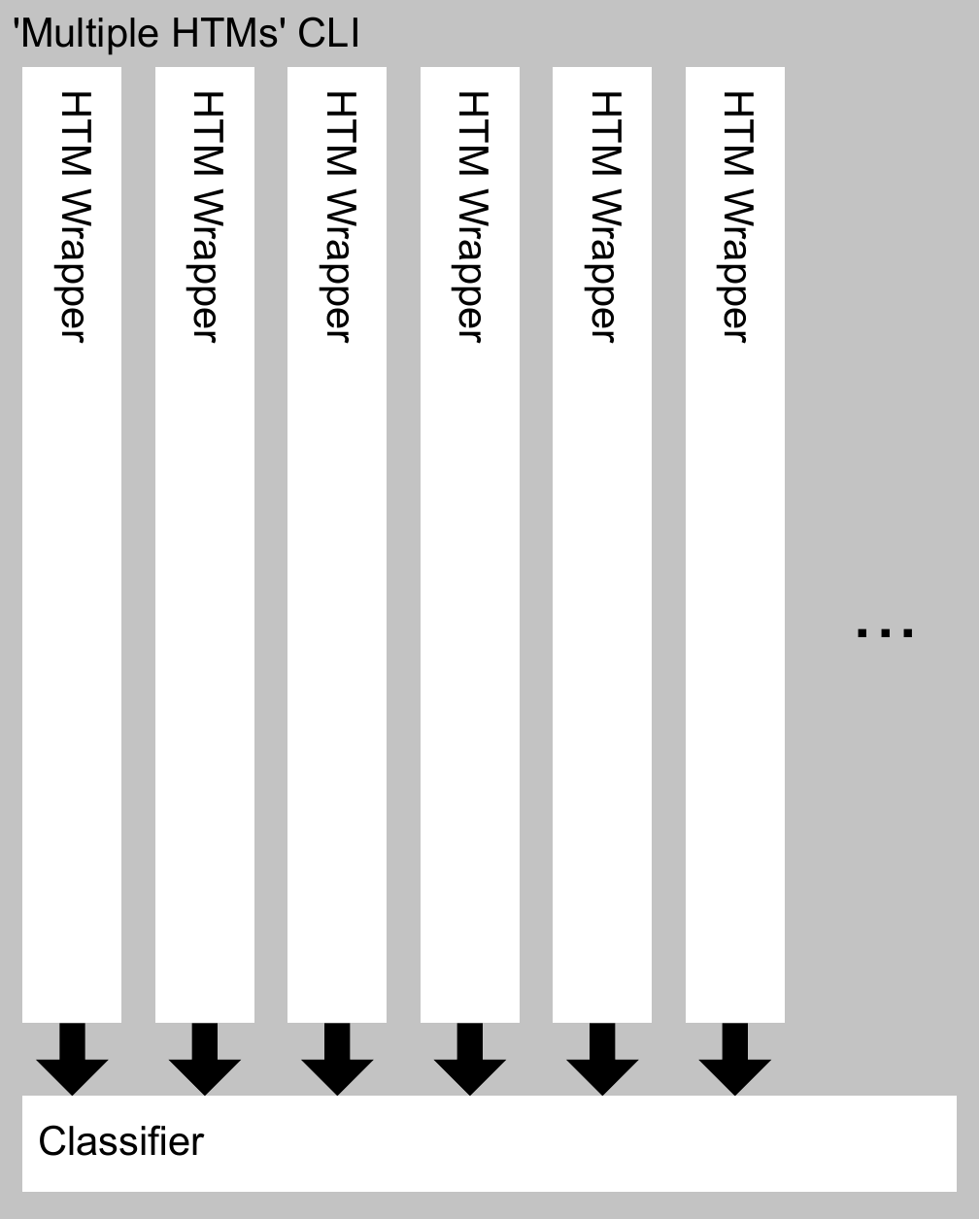}
\end{subfigure}
    \caption{Architecture of the implemented system}
    \label{fig:architecture_of_the_implemented_system}
\end{figure*}

The HTM itself is a 'core' module, in addition to the ones necessary for the system to function (responsible for data reading and encoding, as well as results interpretation) and ones created for debugging and statistics gathering purposes. The overall system architecture is depicted in Fig. \ref{fig:architecture_of_the_implemented_system}. The most relevant modules are described in detail below.

\subsection{Outer Structure}
\label{subsection:outer_structure}

The outermost level of the system is CLI (Command Line Interface). Depending on the provided command line options, it invokes a particular setup -- either 'Single HTM' or 'Multiple HTMs'. In the 'Single HTM' setup, data from all categories are fed into a single HTM instance. 'Multiple HTMs' refers to creating HTM instances on a per--category basis, resulting in an ensemble of one--vs--all detectors.

In both setups the same wrappers encapsulating the actual processing units can be used. A wrapper is created for a particular HTM use -- it is responsible for creating relevant data readers, encoders, decoders and output writers, and for passing them to the iterator -- a part of the core that manages HTM cycles.

After data is processed by the wrapper, the result reaches the CLI, which is responsible for further analysis and data presentation -- combining wrapper outputs, gathering statistics, training the classifier used to provide the final results, rendering data visualizations etc. The HTM results are post-processed using a LinearSVM classifier.

\subsubsection{HTM Wrapper}
\label{subsection:htm_wrapper}

As mentioned above, a wrapper is created for a specific use -- one designed to work with videos will differ from one tailored for texts. Assembling a wrapper from predefined or newly created modules is the main task of the experiment setup.

The wrapper used in the present system setup creates a reader able to get data from video files and an encoder that converts raw frame data to the required format. The HTM output is neither modified (a pass-through decoder module) nor stored for future reference (a pass-through writer module).

Preparing the processing units to work is not the wrapper's only responsibility -- it also controls the number of executed iterations. The minimum (and default) number of cycles equals a single pass of the learning set, however setups specifying a maximum number and/or metrics measuring whether HTM still needs learning are also possible.

The wrapper module also coordinates statistics gathering and visualization on a per-instance basis.

\subsubsection{Adaptive Video Encoder}
\label{subsection:adaptive_video_encoder}

During the encoding process the original video frame is converted to a binary image. Depending on the configuration, the original image can be first reduced in size to trim down the amount of data. After reduction, the color image is converted to a grayscale one, which is later binarized using adaptive thresholding.

Adaptive thresholding uses a potentially different threshold value for each small image region. It gives better results than using a single threshold value for images with varying illumination. In this encoder 'ADAPTIVE\_THRESH\_Gaussian\_C' algorithm from OpenCV library\cite{opencv_library} is used -- a threshold value is the weighted sum of neighboring values where weights are a Gaussian window.

\subsection{HTM Core}
\label{subsection:htm_core}

All implemented readers, encoders, decoders and writers provide pre-defined interfaces. Such a solution allows us to separate data acquisition and output storage from the actual processing. The loop consisting of data retrieval, processing and outputting is executed by the iterator object of the core module.

An HTM object itself consists of a configurable number of layers, a Spatial Pooler and a Temporal Pooler object. Upon each iteration, each layer state is updated by the SP and (depending on the configuration) the TP, based on the data it receives. In the case of the lowest layer the input is obtained from the encoder, and for the higher ones -- from the previous level. Setting the layer number to zero effectively turns off the HTM, causing the whole module's output to be equal to that of the encoder. This feature was used when comparing performance of 'SVM' only with the 'SP + SVM' ensemble.

Layers consist of columns, which are composed of connectors (containing synapses used in the spatial pooling process) and cells (used in temporal pooling). Cells themselves are built from segments, with each segment containing synapses connecting it to the other cells. This hierarchical structure closely mirrors the one described in the algorithm section.

Every object encapsulates its functionality, making introduction of changes and enhancements trivial, while at the same time providing a clear reference point for modifications. The object-oriented structure also enhances the visibility of a very important HTM feature -- its potential for massive parallelization. One example of that can be a spatial pooling process. The initial system setup used a sequential version of SP. After some tests, a decision to replace it with a concurrent implementation running on a GPU (and an FPGA in the future) was made. The replacement spatial pooler, taking advantage of OpenCL capabilities, was written and plugged into the system without changes to the rest of the architecture. The best OpenCL kernel speed-up (running on GPU vs. CPU) of 632x and 207x was reached for 256 synapses and 1024 columns when compared to basic configuration for R4 dataset (see Tab. \ref{tab:exp_parameters}). Overall acceleration including data transfer time between devices amounted to 6.5x and 3.2x respectively. Details of experiments can be found in \cite{wielgosz2016opencl}.

\section{Data flow}
\label{section:processing_flow}

\begin{figure}
\centering
\includegraphics[width=0.45\textwidth]{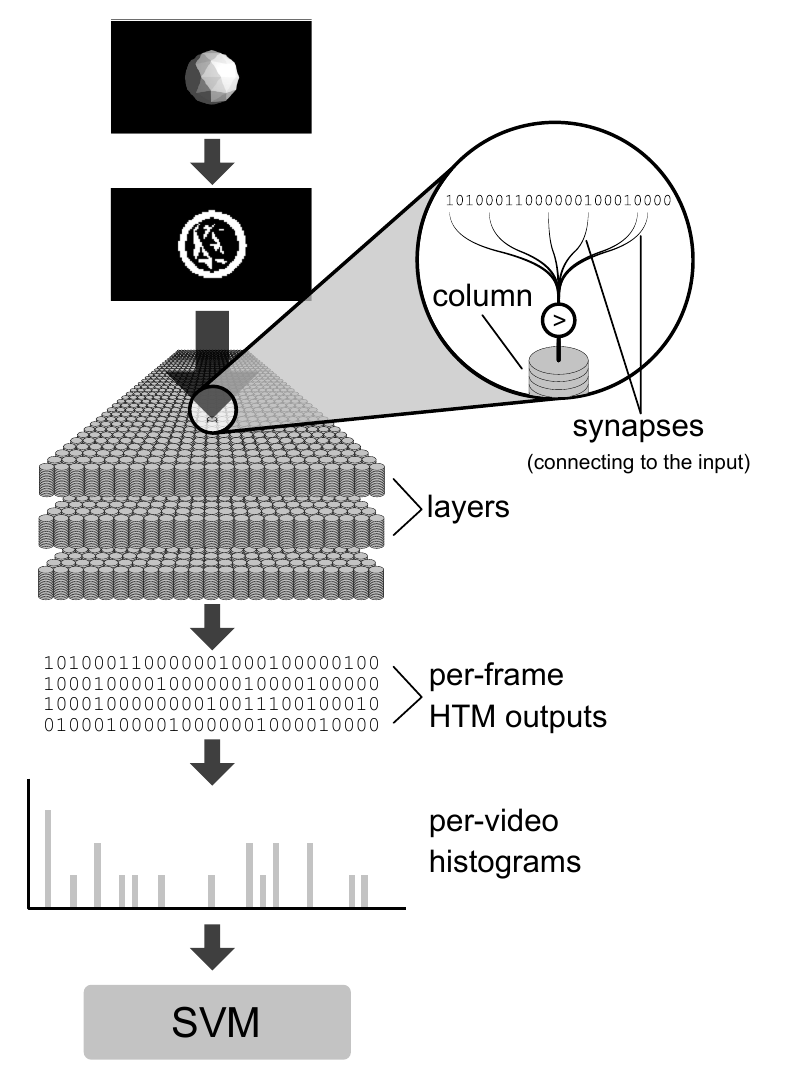}
\caption{Block diagram of the proposed approach}
\label{fig:data_flow}
\end{figure}

The data is fed into the system in a frame--by--frame manner. In the first step, the original frame is turned into a binary image (see \ref{subsection:adaptive_video_encoder}). This conversion constitutes the encoding which allows the generation of input data for the SP processing stage.

Thereafter, the encoded data is fed into the SP. The processing done by the SP effectively maps input to Sparse Distributed Representation (SDR), which then may be passed on to the TP. We do not use TP in this particular application, but the system in general has such a capability. Instead, we substitute TP with histograms to serve a similar purpose.

Histograms of consecutive frames are built from SP output on a per--video basis. When an HTM module is disabled, encoder output is used to create the histograms (see \ref{subsection:htm_core}). They are used as the input data for the SVM classifier which comes next. The classifier maps the results from SDR to the result space (output categories).

The complete processing flow of the system is presented in Fig.~\ref{fig:data_flow}.
\section{Experiments and the discussion}
\label{section:experiments} 

A main goal of the experiments was to examine an ability of the presented model to classify objects in noisy video streams. The authors decided to use artificially generated video sequences for a validation of the presented method. Such an approach allowed to focus on critical aspects of model architecture because the authors could observe the relationships between certain parameters and the quality and complexity of the input data stream. It is worth noting that the Spatial Pooler is unsupervised, so the training process is mostly governed by a proper choice of the model structure. Consequently, during a series of experiments the most important properties of the system were determined. 

In datasets based on real-life scenes it is hard to find and select videos with a well-defined distortions such as insufficient light, certain noise level or a very distant object seen from an unusual angle. Even if such videos exist, tagging them and quantitatively measuring noise or a distortion level is a tedious task. The artificial dataset with complex scene modification capabilities used in experiments poses no such problems. Furthermore, efficient object classification in complex real-life video requires attention mechanism \cite{shikhar2015Action} to be able to find and track right spots in scenes. Therefore, the authors decided to locate figures centrally in a scene.

\subsection{Experiments setup}

\begin{table}
\caption{Experiment parameters}\label{tab:exp_parameters}
\centering
\begin{tabular}{lll}
\toprule
No. of frames per video & \multicolumn{2}{l}{32} \\ \midrule
Object classes & \multicolumn{2}{l}{\makecell[l]{cone, cube, cylinder, \\monkey, sphere, torus}} \\ \midrule
No. of classes & \multicolumn{2}{l}{6} \\ \midrule
\multirow{3}{*}[-0.6em]{Total no. of videos} & all & 6000 \\ \cmidrule{2-3}
 & training & 4800 \\ \cmidrule{2-3}
 & testing & 1200 \\ \midrule
\multirow{3}{*}[-0.6em]{Videos per class} & all & 1000 \\ \cmidrule{2-3}
 & training & 800 \\ \cmidrule{2-3}
 & testing & 200 \\ \midrule
\multirow{3}{*}[-0.6em]{Videos per trial} & all & 100 \\ \cmidrule{2-3}
 & training & 80 \\ \cmidrule{2-3}
 & testing & 20 \\
 \bottomrule
\end{tabular}
\end{table}

\begin{table}
\caption{SP parameters}\label{tab:sp_parameters}
\centering
\begin{tabular}{lccc}
\toprule
 & \multicolumn{3}{c}{Dataset} \\ \cmidrule{2-4}
 & R16 & R8 & R4 \\ \midrule
\makecell[l]{Size of a single\\video frame} & ~60x32~ & 120x66 & 240x134 \\ \midrule
No. of columns & \multicolumn{3}{c}{2048}\\ \midrule
\makecell[l]{No. of synapses\\per column} & 64 & 64 & 128\\ \midrule
\makecell[l]{Perm value\\increment} & \multicolumn{3}{c}{0.1}\\ \midrule
\makecell[l]{Perm value\\decrement} &\multicolumn{3}{c}{0.1}\\ \midrule
Min overlap & \multicolumn{3}{c}{8}\\ \midrule
Winners set size & \multicolumn{3}{c}{40} \\ \midrule
Initial perm value & \multicolumn{3}{c}{0.21} \\ \midrule
\makecell[l]{Initial inhibition\\radius} & \multicolumn{3}{c}{80} \\ \bottomrule
\end{tabular}
\end{table}

A series of experiments (details of which are provided in Tab.~\ref{tab:exp_parameters} and Tab.~\ref{tab:sp_parameters}) was conducted to validate the hypothesis stated in the introduction of the paper. They allow a comparison of the performance of the system featuring Spatial Pooler in the processing flow with the one lacking it (denoted as 'SVM'). Two approaches to introducing SP into the system were tested: 'Single HTM' mode, denoted as 'SP\textsubscript{S} + SVM' and 'Multiple HTMs' mode, denoted as  'SP\textsubscript{M} + SVM'.

\subsection{Debugging}

\begin{figure*}
\centering
\begin{subfigure}{0.78\textwidth}
    \includegraphics[width=\textwidth]{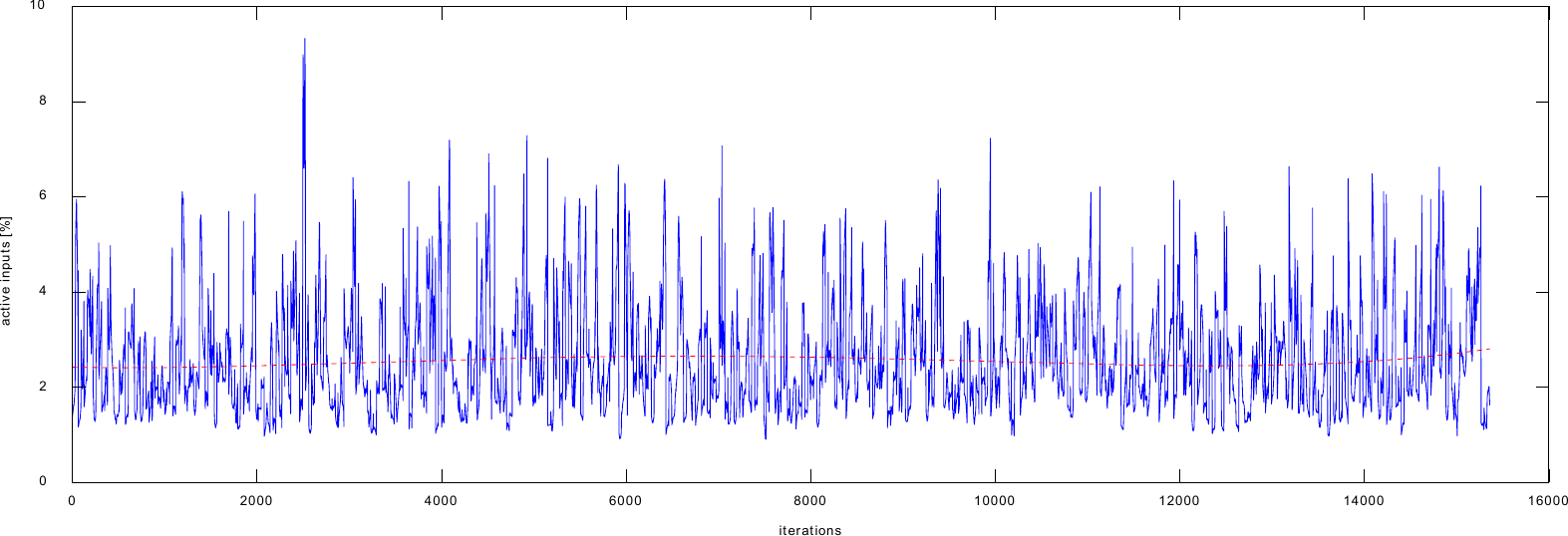}
    \caption{Learning mode}
\end{subfigure}
\begin{subfigure}{0.78\textwidth}
    \vspace{6pt}
    \includegraphics[width=\textwidth]{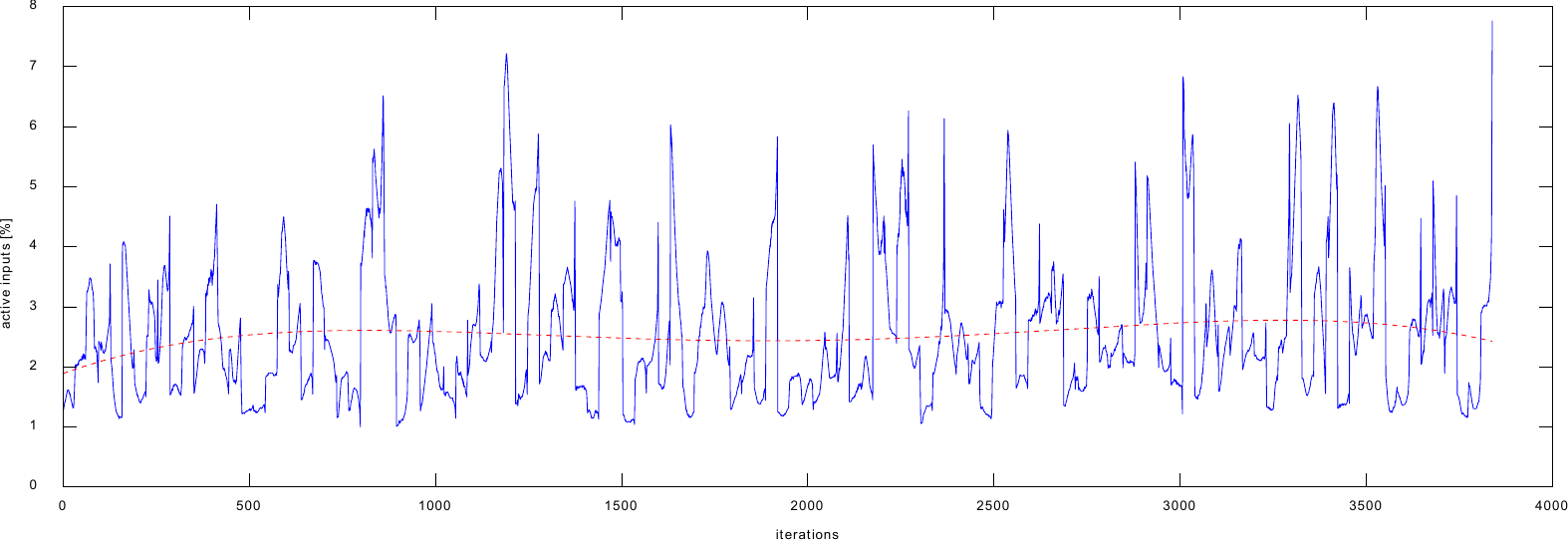}
    \caption{Testing mode}
\end{subfigure}
\caption{Sample visualizer plots depicting percentage of active bits in an input frame.}
\label{fig:visualizer_active_inputs}
\end{figure*}

\begin{figure*}
\centering
\begin{subfigure}{0.78\textwidth}
    \includegraphics[width=\textwidth]{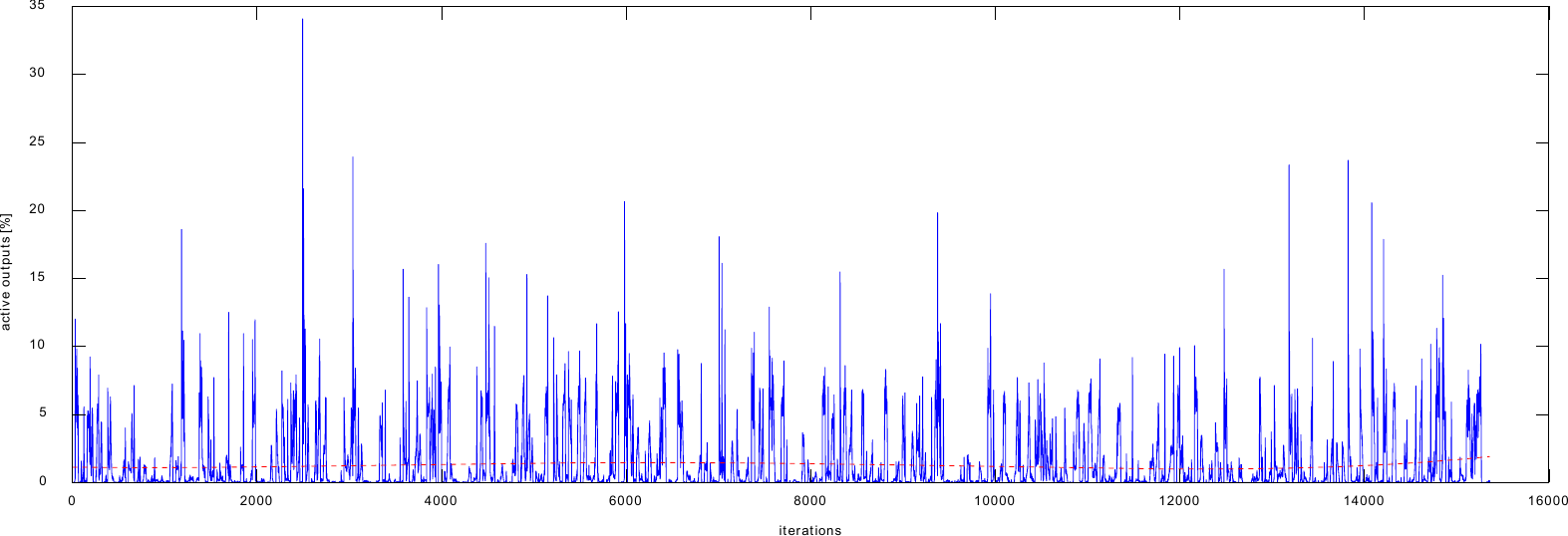}
    \caption{Learning mode}
\end{subfigure}
\begin{subfigure}{0.78\textwidth}
    \vspace{6pt}
    \includegraphics[width=\textwidth]{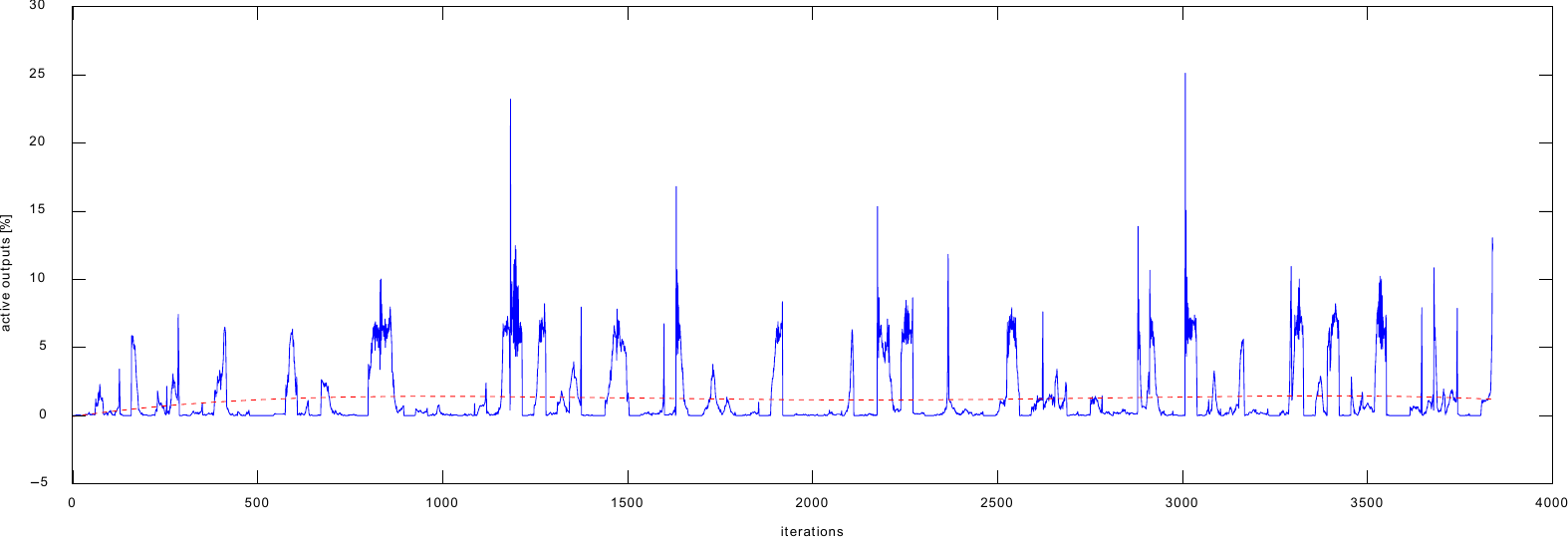}
    \caption{Testing mode}
\end{subfigure}
\caption{Sample visualizer plots depicting percentage of active HTM outputs.}
\label{fig:visualizer_active_outputs}
\end{figure*}

\begin{figure*}
\centering
\begin{subfigure}{0.78\textwidth}
    \includegraphics[width=\textwidth]{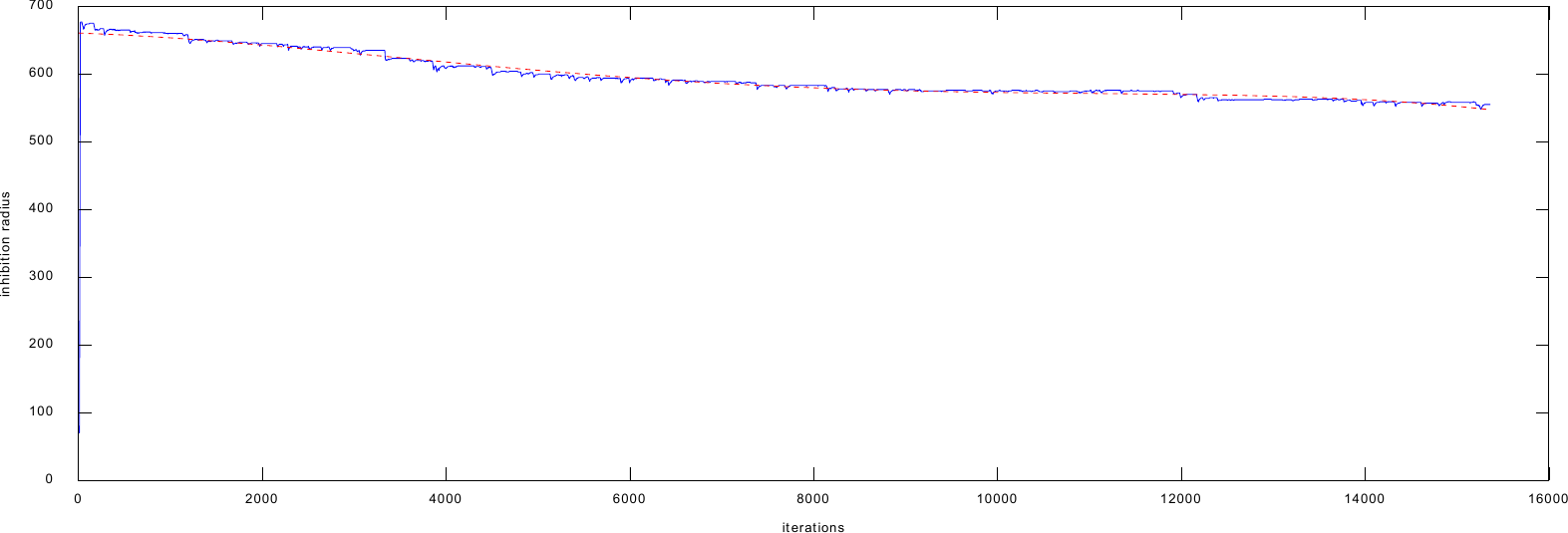}
    \caption{Learning mode}
\end{subfigure}
\begin{subfigure}{0.78\textwidth}
    \vspace{6pt}
    \includegraphics[width=\textwidth]{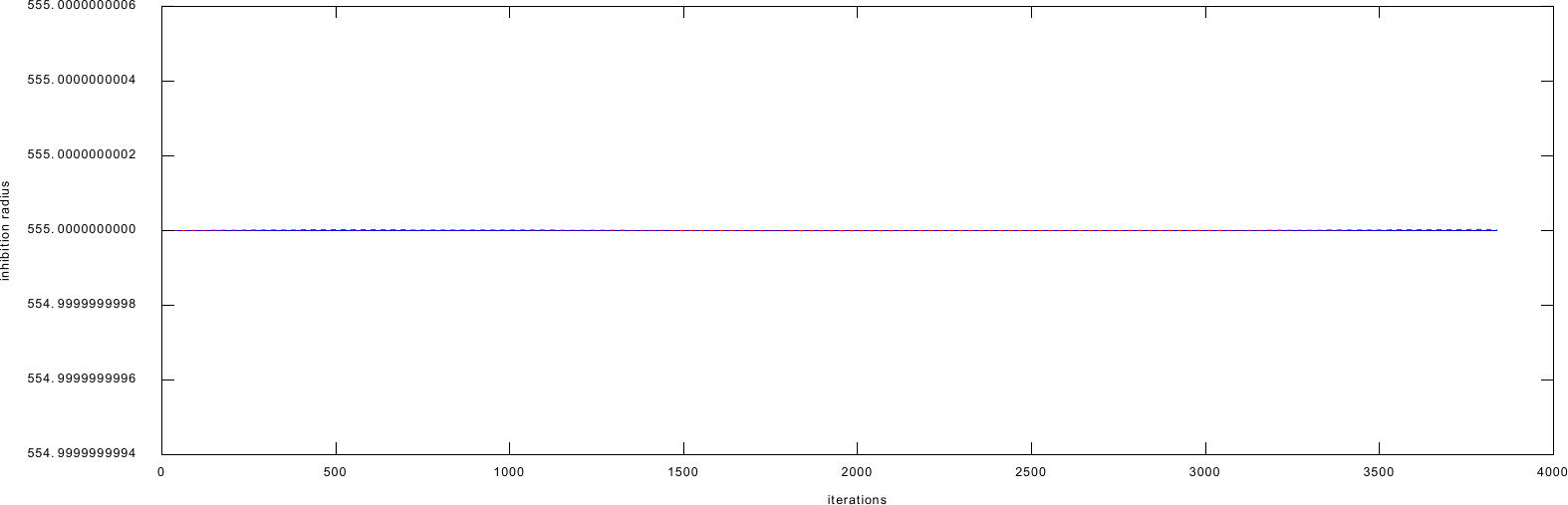}
    \caption{Testing mode}
\end{subfigure}
\caption{Sample visualizer plots depicting changes in inhibition radius value.}
\label{fig:visualizer_inhibition_radius}
\end{figure*}

\begin{figure}
\includegraphics[width=0.48\textwidth]{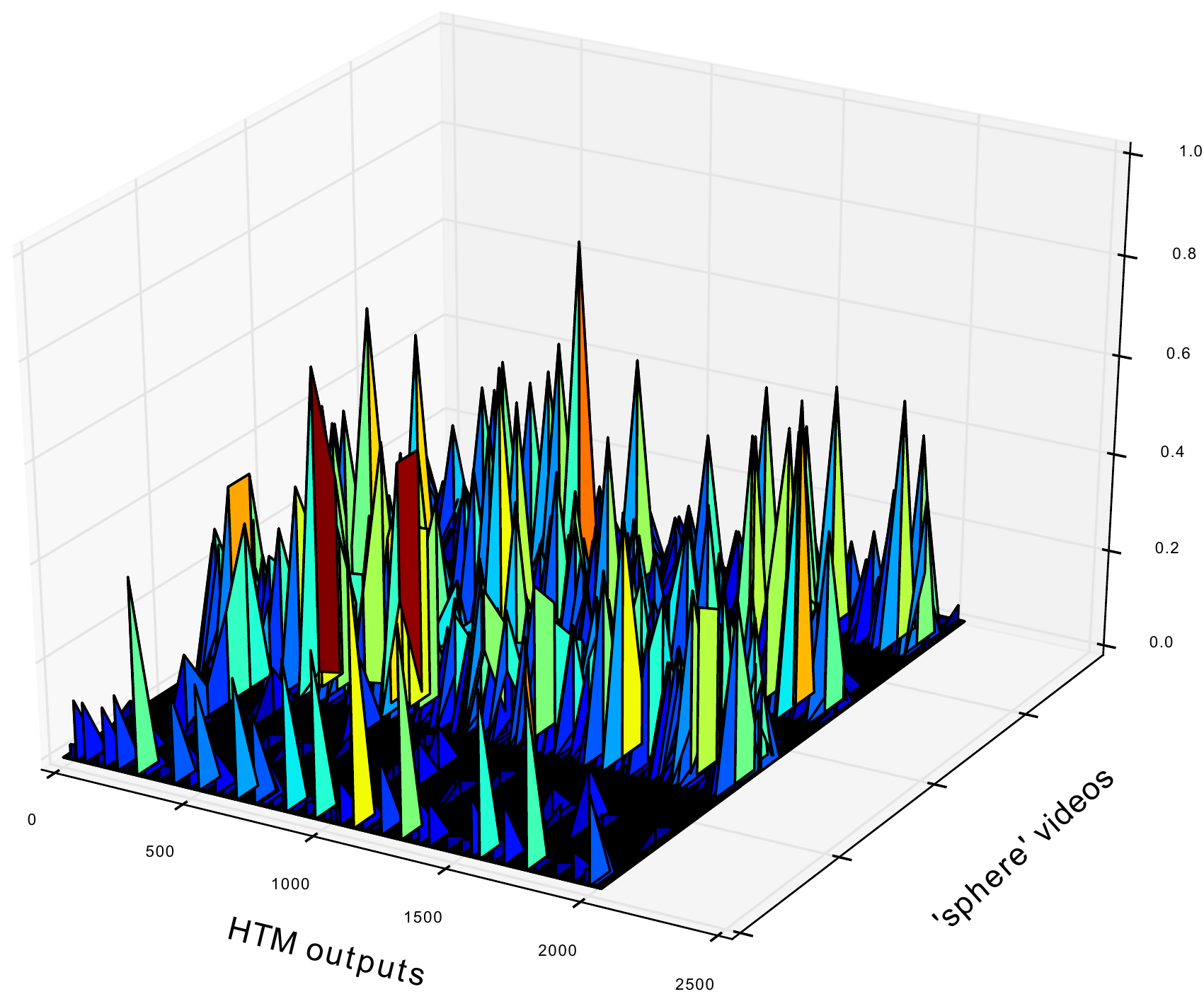}
\caption{Sample visualizer plots depicting class histograms for videos belonging to the 'sphere' class.}
\label{fig:visualizer_histograms}
\end{figure}

Debugging a conventional system is a relatively simple and well-established practice. Challenges arise when addressing other than the von Neumann architectures, such as HTM. In those cases, it is necessary to use unconventional methods and tools, often specifically designed to debug and analyze the particular application. The authors have faced the same challenges during the early stages of the system development. The initially designed tool that allowed them to analyze individual columns and synapses did not allow for effective tracing of the module behavior, because the system as a whole behaves statistically. Therefore, the authors decided to develop a tool that can globally profile the whole HTM system. It consists of two modules: an analyzer and a visualizer. The first one stores system operating parameters (e.g. active inputs, inhibition radius), while the second is responsible for data presentation in a form easily understandable to the designer of the HTM system. The debugging methods and tools are part of the system available at \cite{online:custom_htm} and may be used in experiments. They allow to specify a set of parameters which are to be included in a final report. Sample visualizations are shown in Fig.~\ref{fig:visualizer_active_inputs} -- \ref{fig:visualizer_histograms}.

It is worth emphasizing that the tool contributed a lot to the development of the presented system and in the authors opinion it would have been hard to bring the module the its final shape without it. Columns and synapses number adjusting may be given as an example of the challenge the authors managed to address with the tool. During the development the system yielded wrong results, classifying nearly all videos to a single class. It was hard to determine what was the cause, but after a close examination it turned out that input data coverage was insufficient. This in turn was caused by the low number of synapses per column. The authors were able to find a cause by scrutinizing a plot of the overlap value generated by the visualizer. It turned out that for a certain input data overlap dropped to zero despite the fact that those data items had been presented to the network in a training process before. In the next step, a number of synapses was increased which alleviated the problem by reducing a number of wrongly classified items. The presented verification and reasoning flow was substantially based on the debugging.

It is worth noting that despite that fact that substantial section of the source code of the system were ported to GPU for better performance, in the debugging mode they will be run on CPU. This affects the simulation time significantly and should be taken into account.

When the system or HTM Core \cite{online:custom_htm} does not work properly, it is recommended to use the debugging tool. According the authors opinion, the best approach is to start with most critical parameters of the network, especially when a designer has no track or clue what may be wrong. Gradually, by examining the plots generated for the input, column and output activity, as well as overlap changes, one can deduct about the condition of the system. Further steps may involve picking some other parameters, such as the inhibition radius or boost, when it turns out that they may play a substantial role in the system behavior.

\subsection{Datasets}

\begin{figure*}
\includegraphics[width=\textwidth]{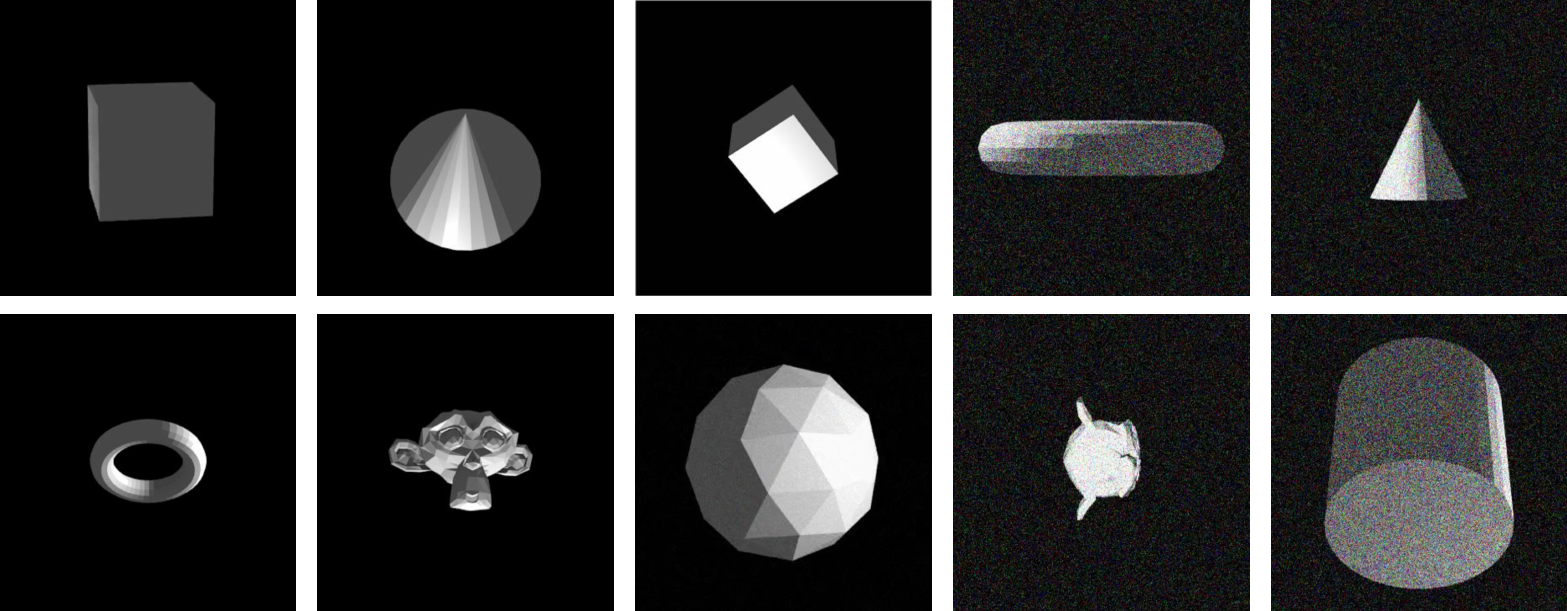}
\caption{Sample (cropped) frames of different shapes and noise levels.}
\label{fig:video_screenshots}
\end{figure*}

\begin{figure}
\includegraphics[width=0.48\textwidth]{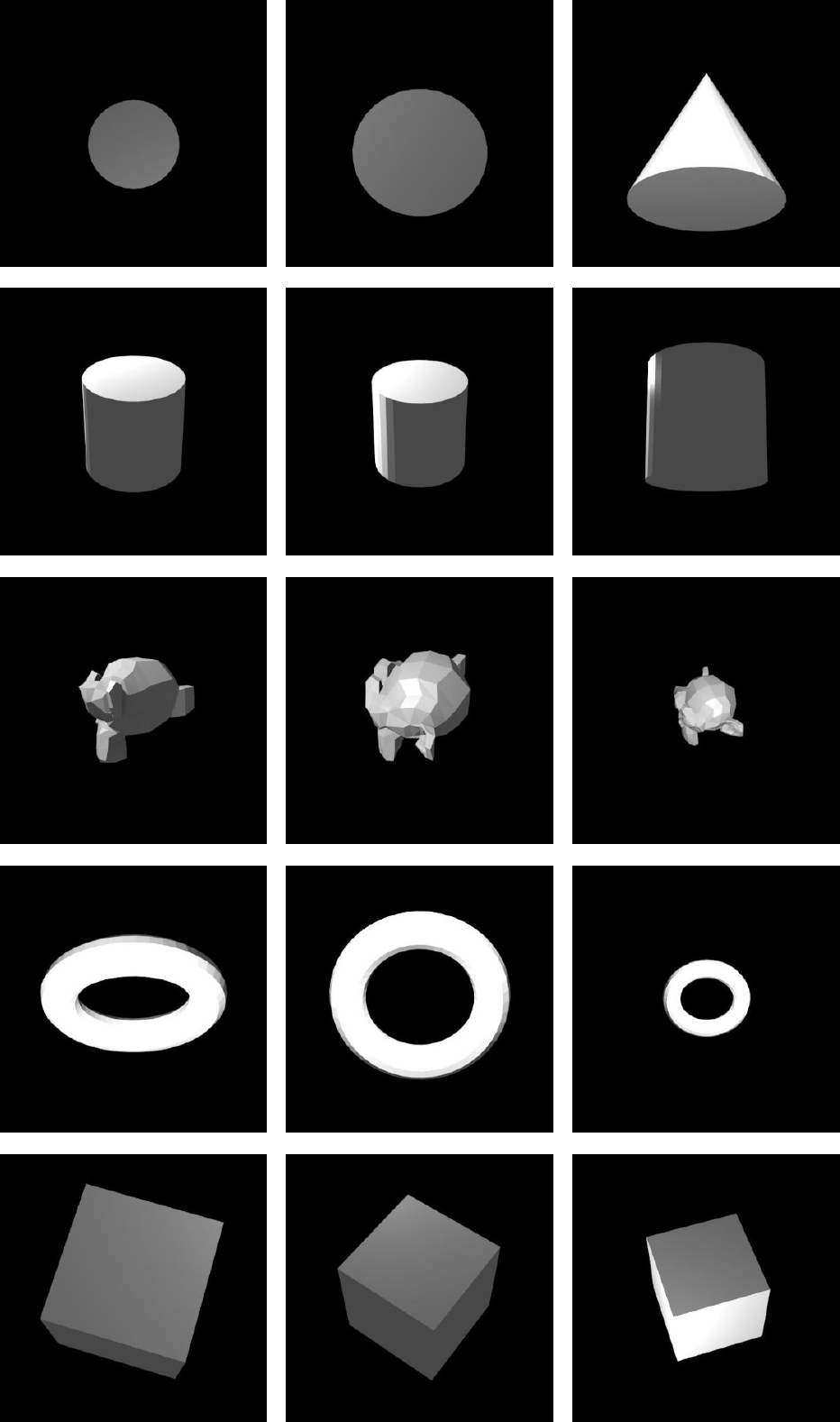}
\caption{The beginning, middle and end of sample videos.}
\label{fig:video_movement_short}
\end{figure}

The challenging part involved generation of sample videos for testing. The videos had to meet a series of requirements such as object location, camera location and object--camera distance. Consequently, a dedicated application was used to generate the videos (i.e. Blender \cite{blender}). Blender provides Python API, which allowed authors to automate and randomize the video generation. Consequently, a series of videos were generated of 960x540 pixels each. All the footages contain a single, centered, stationary object with the camera moving around it in all directions and at varying distance. The sample frames are presented in Fig \ref{fig:video_screenshots} and \ref{fig:video_movement_short}. 

For the experiments, three datasets (available online \cite{online:datasets}) based on the rendered videos were created: R16, R8 and R4, with frame sizes  60x32, 120x66 and 240x134 respectively. 
Subsequently, a Gaussian noise was introduced to the testing videos. Since noise addition at the runtime proved to be a very time consuming process, a separate script introducing noise to a large set of generated videos was created.

\subsection{Quality evaluation measure}

The F1 score is used as a quality evaluation of the experiments' results presented in this paper.  
The precision and recall for corresponding clusters are calculated as follows:
\begin{equation}
 Recall(i, j) = \frac{n_{ij}}{n_i},
 \label{eq:recallClustering}
\end{equation}
\begin{equation}
 Precision(i, j) = \frac{n_{ij}}{n_j},
 \label{eq:precisionClustering}
\end{equation}
where $n_{ij}$ is the number of items of class \textit{i} that are classified as members of cluster \textit{j}, while $n_j$ and $n_i$ are the numbers of items in cluster \textit{j} and class \textit{i}, respectively.
The cluster's F1 score is given by the following formula:
\begin{equation}
 F(i, j) = 2 \cdot \frac{Recall(i, j)Precision(i, j)}{Precision(i, j) + Recall(i,j)}.
 \label{eq:FractionalFmeasure}
\end{equation}
The overall quality of the classification can be obtained by taking the weighted average F1 scores for each class. It is given by the equation:
\begin{equation}
 F1 = \sum\limits_{i} \frac{n_i}{n}max{F(i,j)},
 \label{eq:OverallFmeasure}
\end{equation}
where the maximum is taken over all clusters and \textit{n} is the number of all objects. The F1 score value ranges from 0 to 1, with the higher value indicating the higher clustering quality.

\subsection{Results}

\begin{table}
\caption{Experiments results for reduction rate = 4}\label{tab:experiments_results}
\centering
\begin{tabular}{cccc}
\toprule
\multirow{2}{*}[-0.3em]{\makecell{Noise level\\($\sigma$)}} & \multicolumn{3}{c}{F1 score} \\ \cmidrule{2-4}
 & SVM & SP\textsubscript{S} + SVM & SP\textsubscript{M} + SVM \\ \midrule
0 & 0.78 & 0.76 & 0.87 \\ 
4.25 & 0.78  &  0.74 & 0.87 \\ 
8.5 & 0.77 & 0.74 & 0.86 \\
\bottomrule
\end{tabular}
\end{table}

\begin{table}
\caption{Best results obtained for reduction rate = 4 and $\sigma = 0$}\label{tab:best_results}
\centering
\begin{tabular}{lcccc}
\toprule
\multirow{2}{*}[-0.3em]{Parameter} & \multicolumn{2}{c}{SP\textsubscript{S} + SVM} & \multicolumn{2}{c}{SP\textsubscript{M} + SVM} \\ \cmidrule{2-5}
 & \small{Value} & \small{F1 score} & \small{Value} & \small{F1 score} \\ \midrule
Columns & 4096 &  0.81 & 4096 & 0.89 \\ 
Synapses & 256 & 0.92 & 256 & 0.94 \\
Min overlap & 4 & 0.95  & 6 & 0.96 \\ 
\makecell[l]{Winners\\set size} & 12 & 0.81 & 28 & 0.89 \\
\bottomrule
\end{tabular}
\end{table}

Tab. \ref{tab:experiments_results} shows performance results of SVM and both singular and modular SP versions with various noise levels and reduction rate fixed to 4. The rest of the parameters of the setup used for the tests were of a standard value as presented in Tab. \ref{tab:sp_parameters}. It should be noted that the 'Multiple HTMs' setup outperforms the 'Singular HTM' one.
The best results achieved for different sets of parameters were included in Tab. \ref{tab:best_results} in order to show the maximum capabilities of the examined setup. 

\begin{figure}
\begin{subfigure}{0.48\textwidth}
    \hspace{-12pt}
    \includegraphics[width=\textwidth]{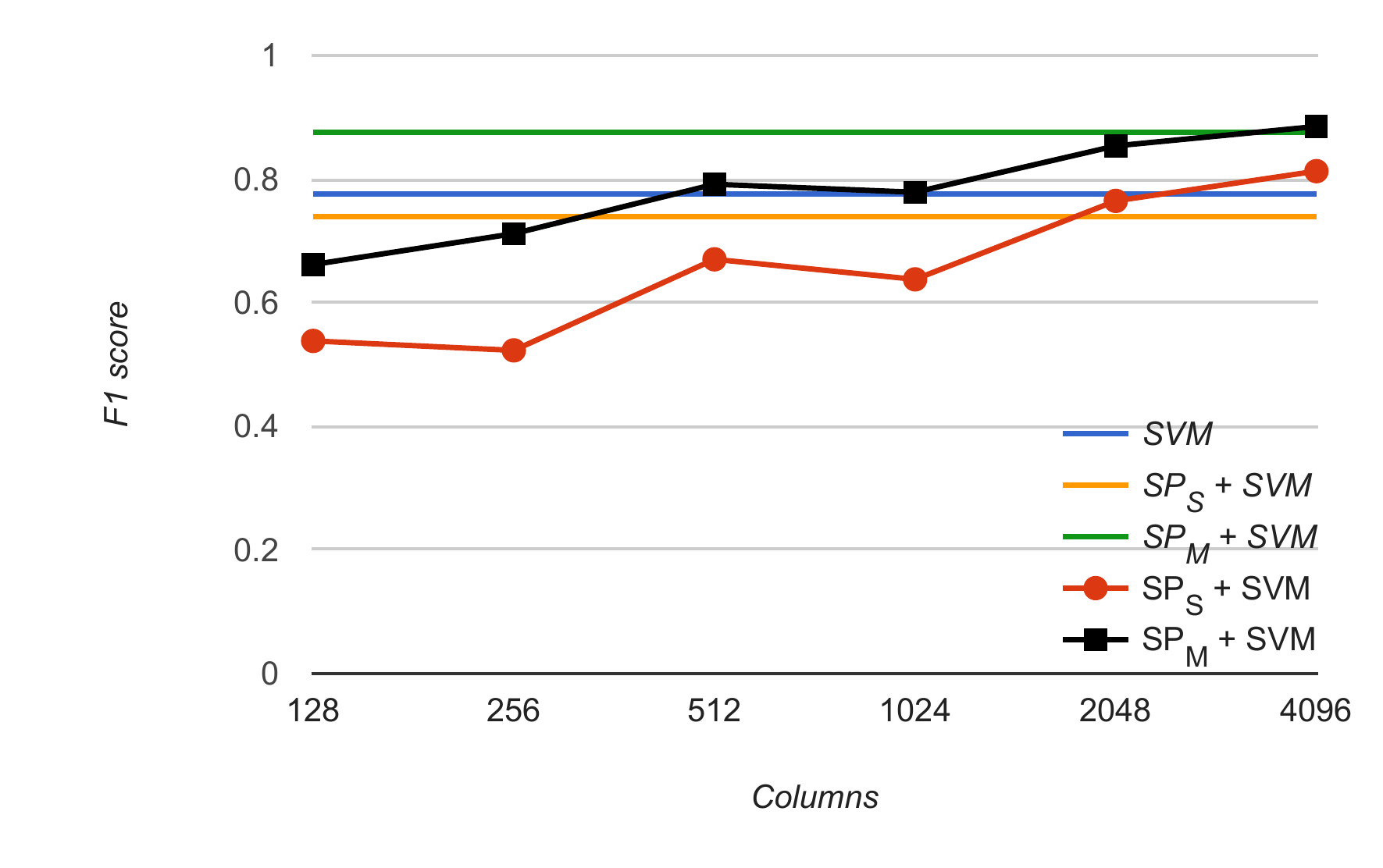}
    \vspace{-10pt}
    \caption{}
\end{subfigure}
\begin{subfigure}{0.48\textwidth}
    \hspace{-12pt}
    \includegraphics[width=\textwidth]{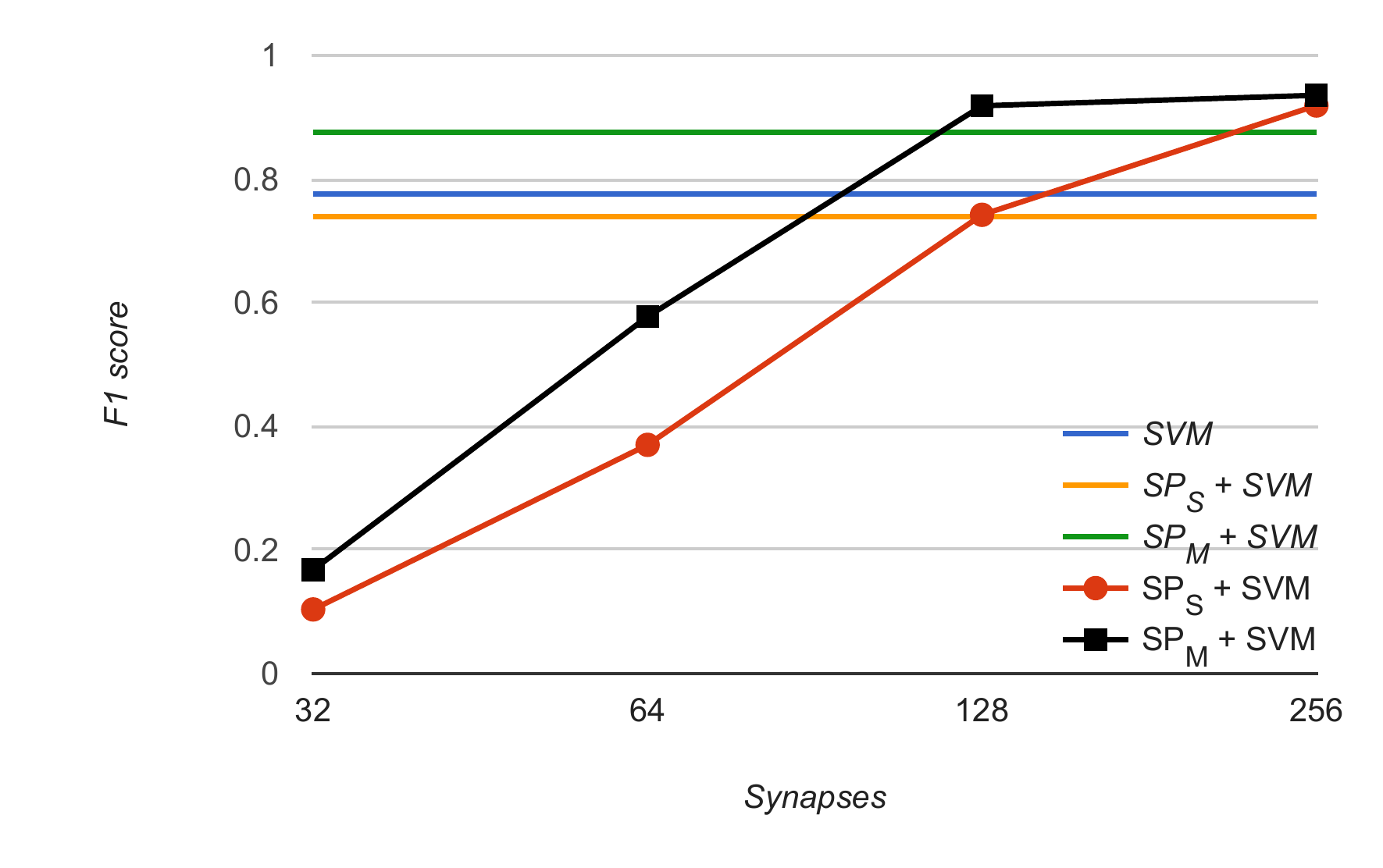}
    \vspace{-10pt}
    \caption{}
\end{subfigure}
\begin{subfigure}{0.48\textwidth}
    \hspace{-12pt}
    \includegraphics[width=\textwidth]{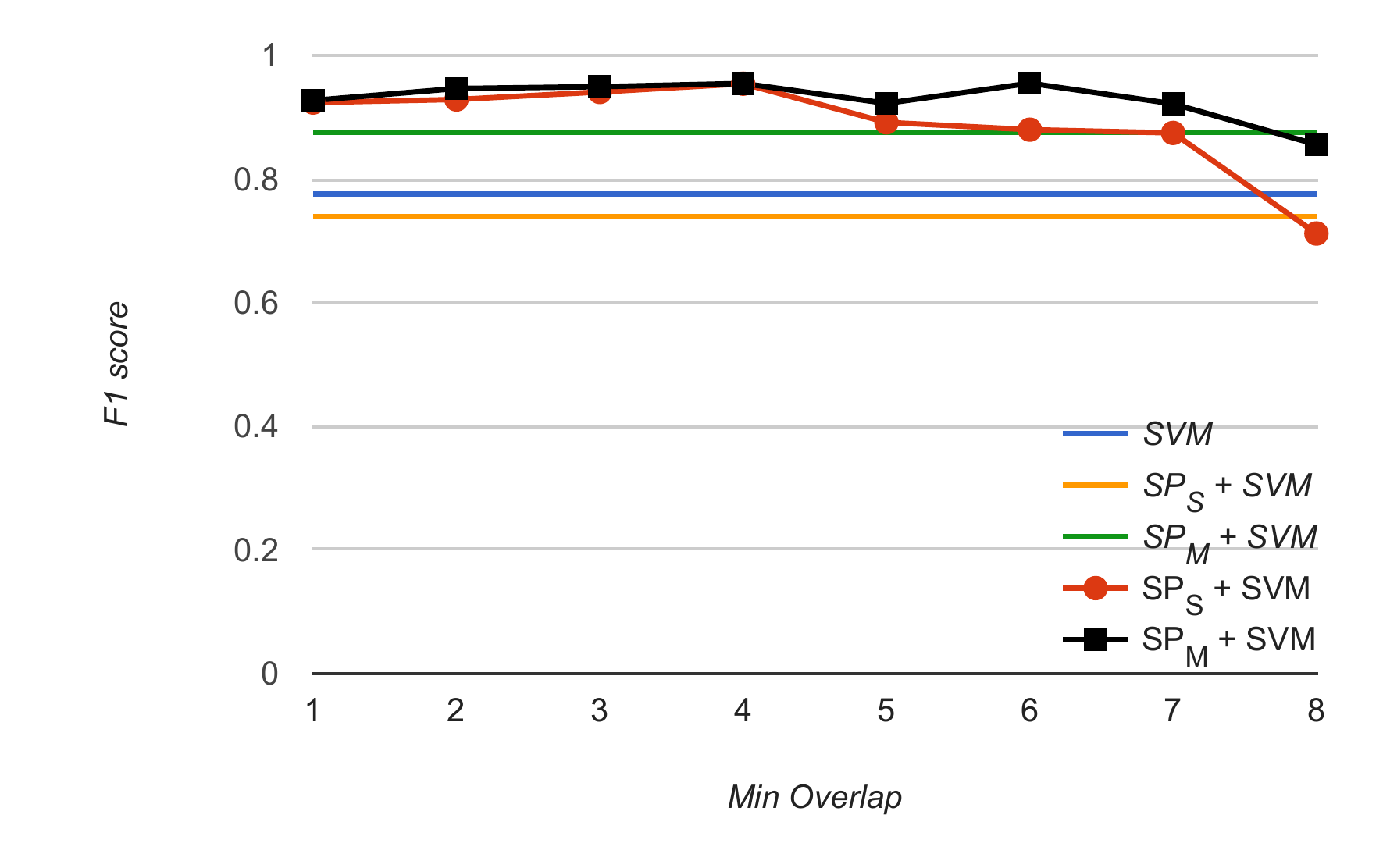}
    \vspace{-10pt}
    \caption{}
\end{subfigure}
\begin{subfigure}{0.48\textwidth}
    \hspace{-12pt}
    \includegraphics[width=\textwidth]{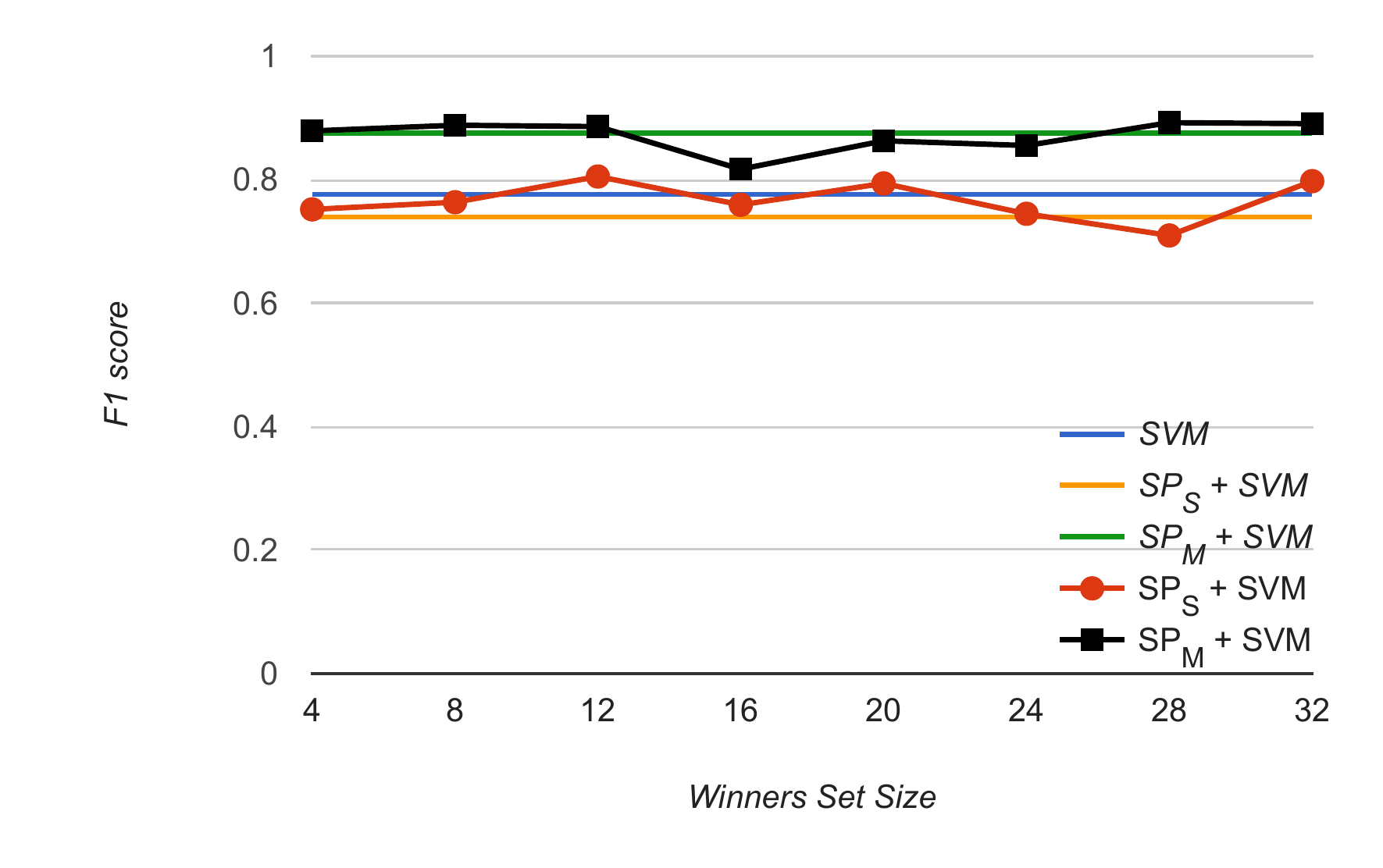}
    \vspace{-10pt}
    \caption{}
\end{subfigure}
\caption{F1 score as a function of different configuration parameters (using R4 dataset). \textit{Italicized labels} denote reference lines calculated for basic configuration.}\label{fig:f1_params}
\end{figure}

In some cases, the results of both setups are almost equal, although 'Multiple HTMs' is slightly superior. Analyzing the results shown in Tab. 4 one easily notices that the highest F1 scores are reached for 256 and 4096 of synapses and columns respectively. Value of $min\_overlap$ also affects the quality of the results and in the case of this setup $min\_overlap=4$ achieves the best performance. Both Tab. \ref{tab:experiments_results} and Fig. \ref{fig:f1_params} show that $winners\_set\_size$ has negligible effect on the overall F1 score.

\begin{table}
\caption{$\dfrac{\text{SP\textsubscript{S} + SVM}}{\text{SVM}}$  histograms cosine similarity ratio}\label{tab:similarity_ratio}
\centering
\begin{tabular}{lcc}
\toprule
\multirow{2}{*}[-1em]{Class} & \multicolumn{2}{c}{Cosine similarity ratio [x]} \\ \cmidrule{2-3}
& \makecell[c]{$\sigma = 4.25$ \\ $\approx 13\%$ of noise bits} & \makecell[c]{$\sigma = 8.5$ \\ $\approx 24\%$ of noise bits} \\ \midrule
cone & 11.56 & 3.54\\ 
cube & 18.14 & 4.27\\ 
cylinder & 13.49 & 4.86\\ 
monkey & 8.19 & 3.16\\ 
sphere & 11.69 & 3.54\\ 
torus & 11.85 & 2.77\\ \midrule
all & 12.64 & 3.73\\
\bottomrule
\end{tabular}
\end{table}

Tab. \ref{tab:similarity_ratio} shows how efficient the SP module is in noise reduction. The values ​​in the columns of the table represent the cosine similarity ratio of the signal before and after SP processing for all the video categories and their average value. 
Cosine similarities were calculated between video histograms yielded by the system for clean video ($\sigma=0$) and the ones with Gaussian noise introduced. Average values were calculated for each class separately as well as for all the videos combined, discarding samples for which cosine similarity could not be calculated.
It should be noted that with increasing levels of noise, SP filtration efficiency decreases, which is expected. In order to increase or maintain its previous efficiency of noise reduction, some of the macro parameters of the system should be tuned to increase selectivity of the module and boost noise elimination.

\begin{figure}
\begin{subfigure}{0.48\textwidth}
    \hspace{-12pt}
    \includegraphics[width=\textwidth]{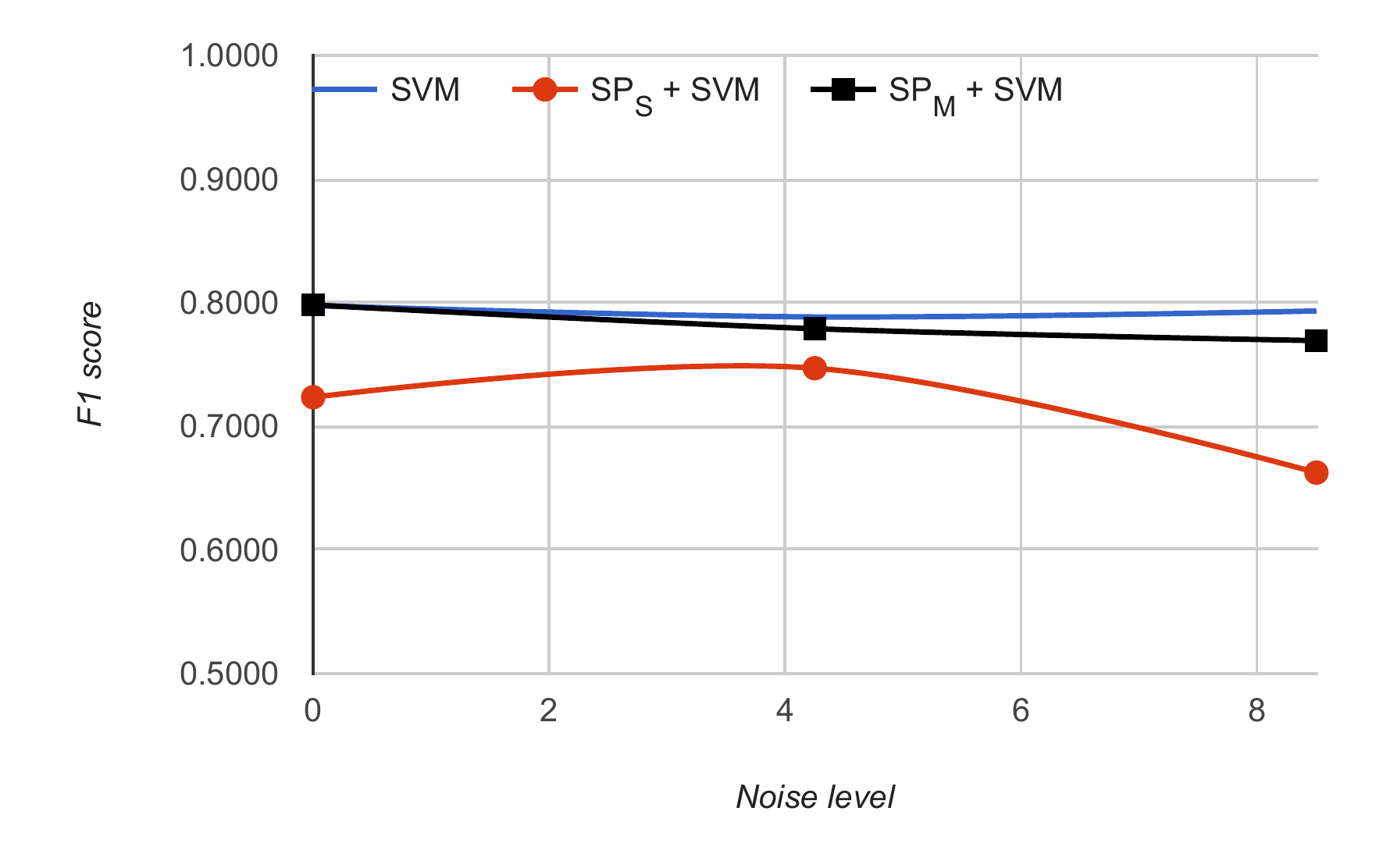}
    \vspace{-8pt}
    \caption{Reduction rate = 16}
\end{subfigure}
\begin{subfigure}{0.48\textwidth}
    \hspace{-12pt}
    \includegraphics[width=\textwidth]{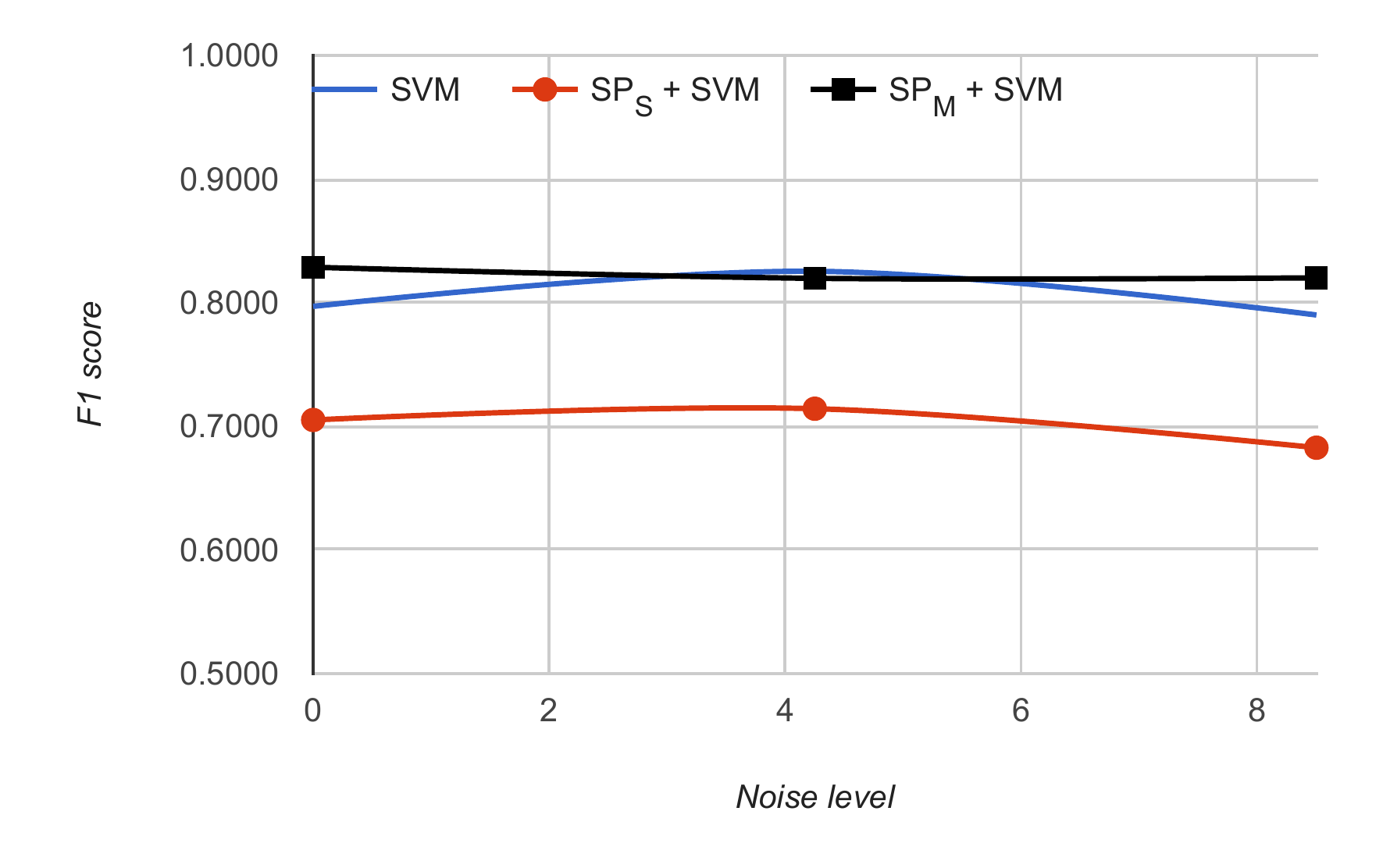}
    \vspace{-8pt}
    \caption{Reduction rate = 8}
\end{subfigure}
\begin{subfigure}{0.48\textwidth}
    \hspace{-12pt}
    \includegraphics[width=\textwidth]{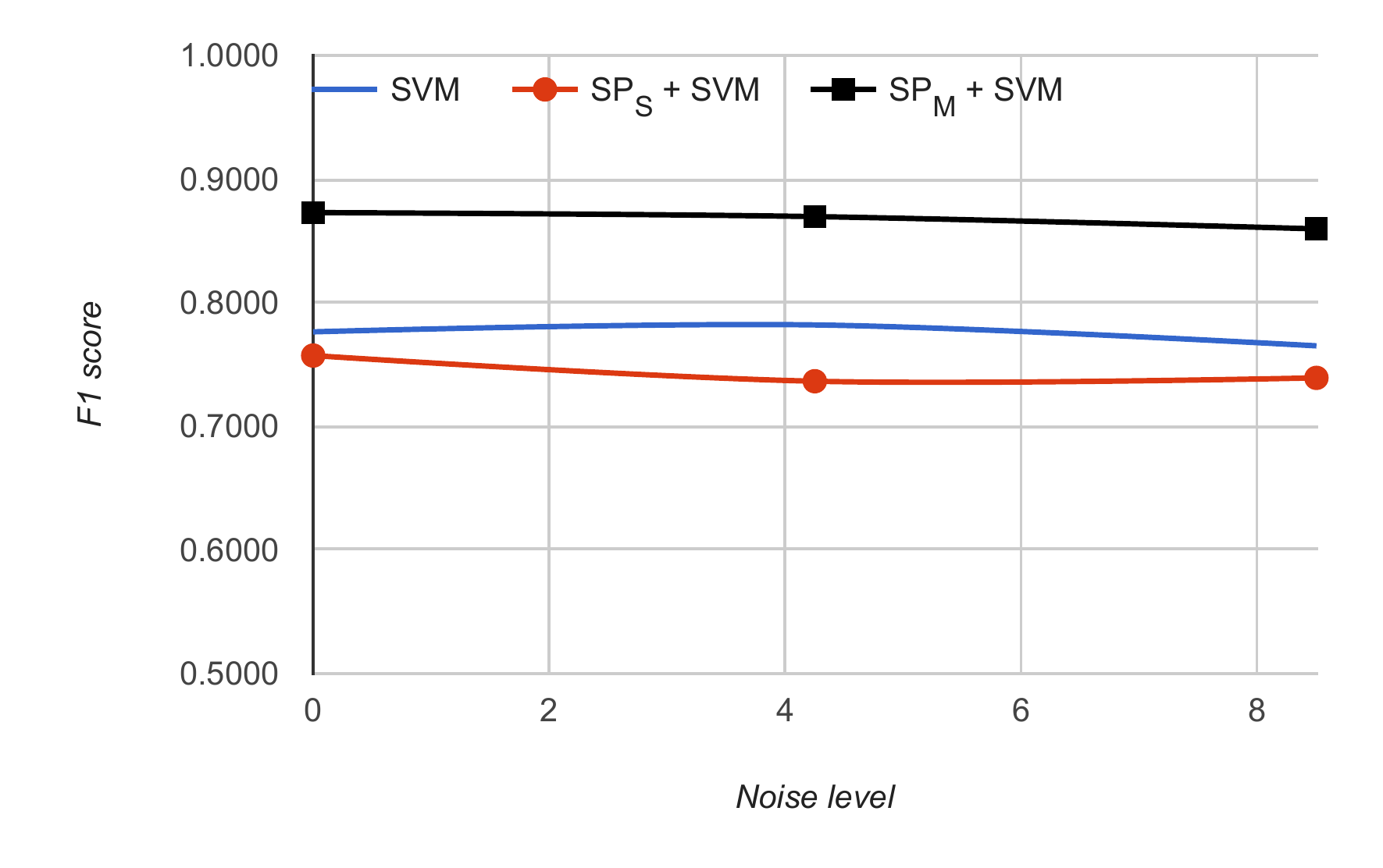}
    \vspace{-8pt}
    \caption{Reduction rate = 4}
\end{subfigure}
\begin{subfigure}{0.48\textwidth}
    \hspace{-12pt}
    \includegraphics[width=\textwidth]{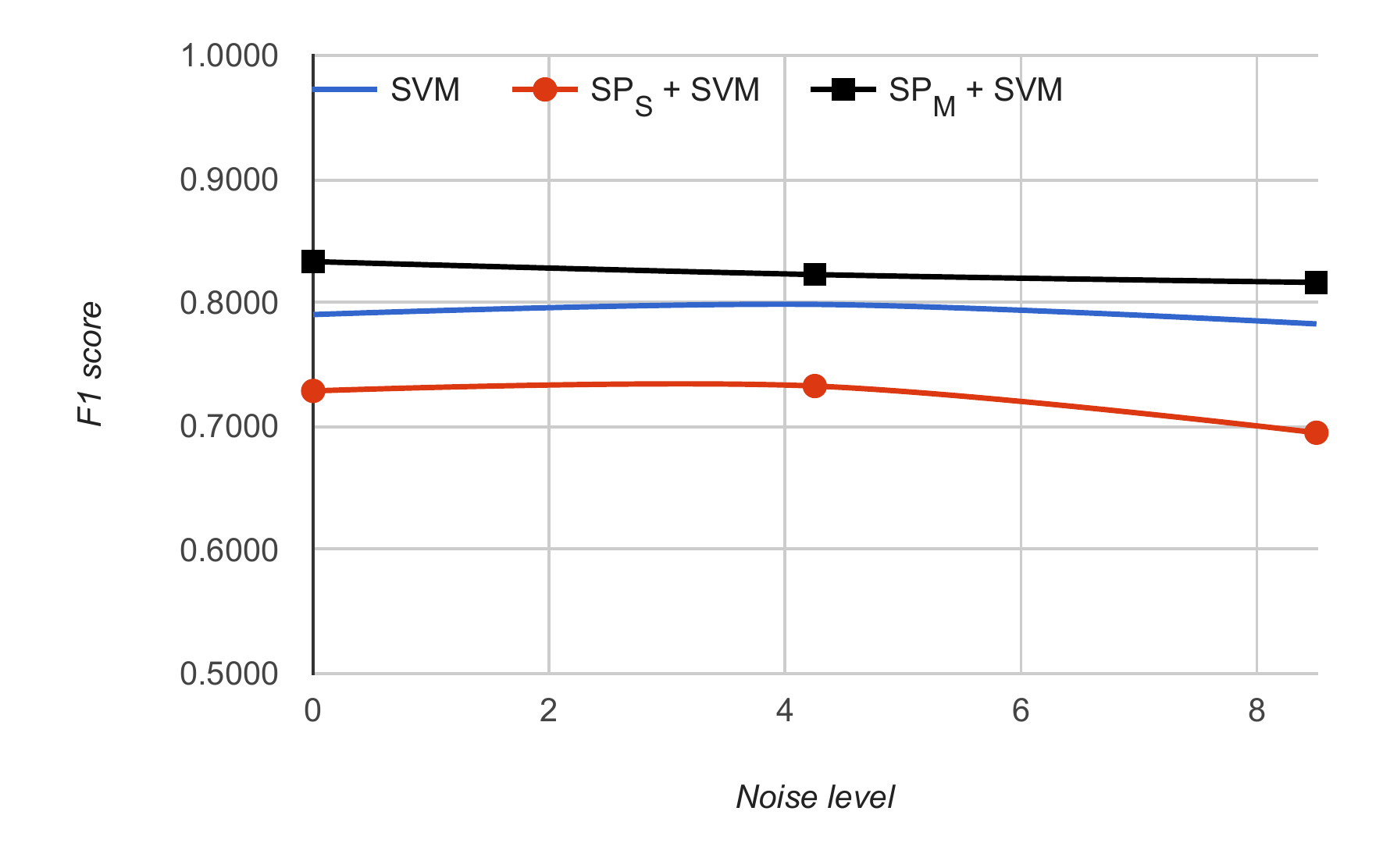}
    \vspace{-8pt}
    \caption{Average}
\end{subfigure}
\caption{F1 score for different noise levels and reduction ratio.}
\label{fig:f1_noise}
\end{figure}

Fig. \ref{fig:f1_noise} presents performance in terms of F1 score for different noise levels. It is worth noting the performance of the system depends both on noise level and reduction rate. As expected, a larger reduction ratio degrades performance of SP + SVM because less data is available. Furthermore, image preprocessing operations and especially binarization introduces distortions which are proportional to the noise level. SP + SVM setup is particularly sensitive to relocation distortions i.e. changing spatial positions of pixels in the data fed into the module.

\begin{figure}
\begin{subfigure}{0.48\textwidth}
    \hspace{-12pt}
    \includegraphics[width=\textwidth]{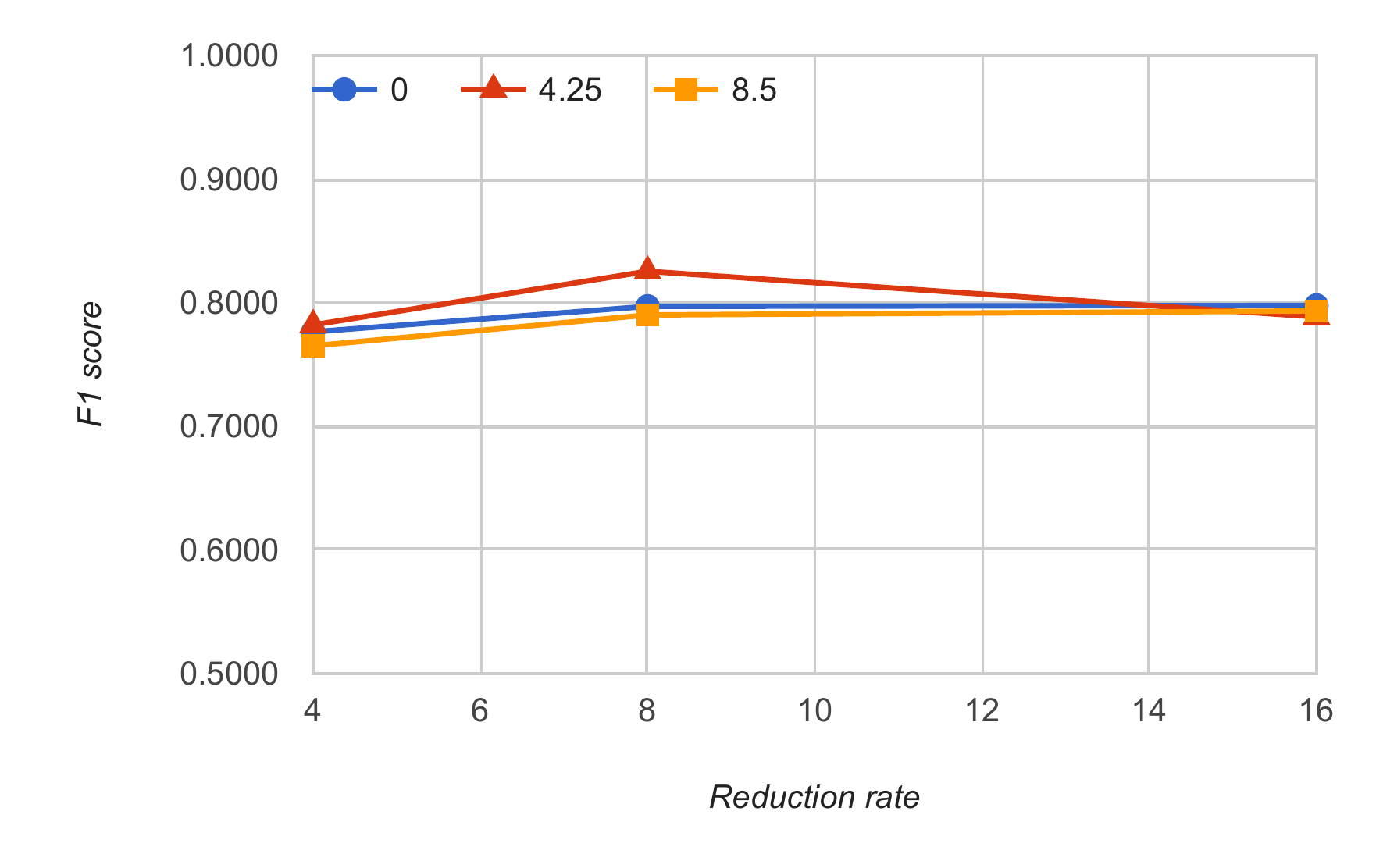}
    \vspace{-8pt}
    \caption{SVM}
\end{subfigure}
\begin{subfigure}{0.48\textwidth}
    \hspace{-12pt}
    \includegraphics[width=\textwidth]{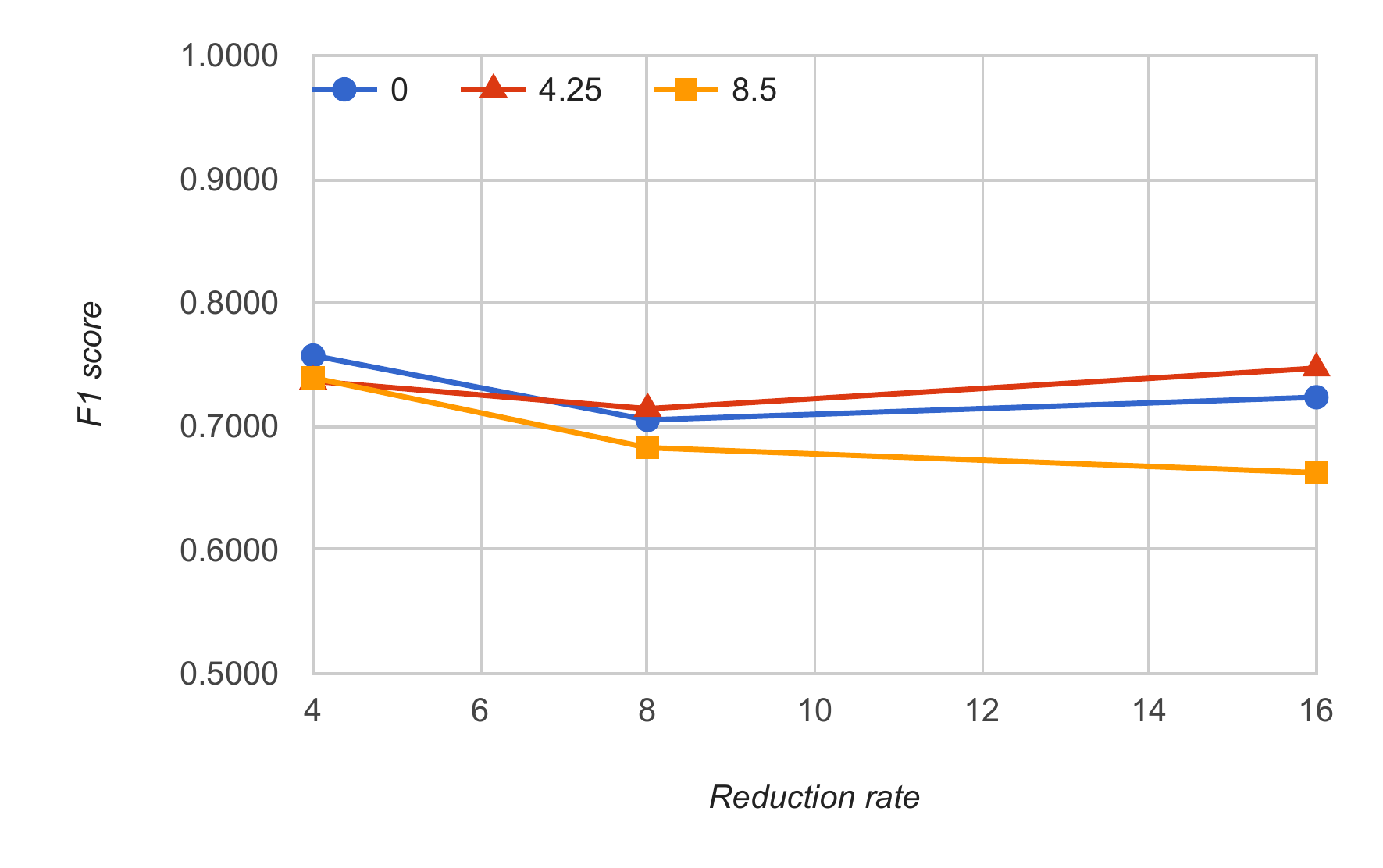}
    \vspace{-8pt}
    \caption{SP\textsubscript{S} + SVM}
\end{subfigure}
\begin{subfigure}{0.48\textwidth}
    \hspace{-12pt}
    \includegraphics[width=\textwidth]{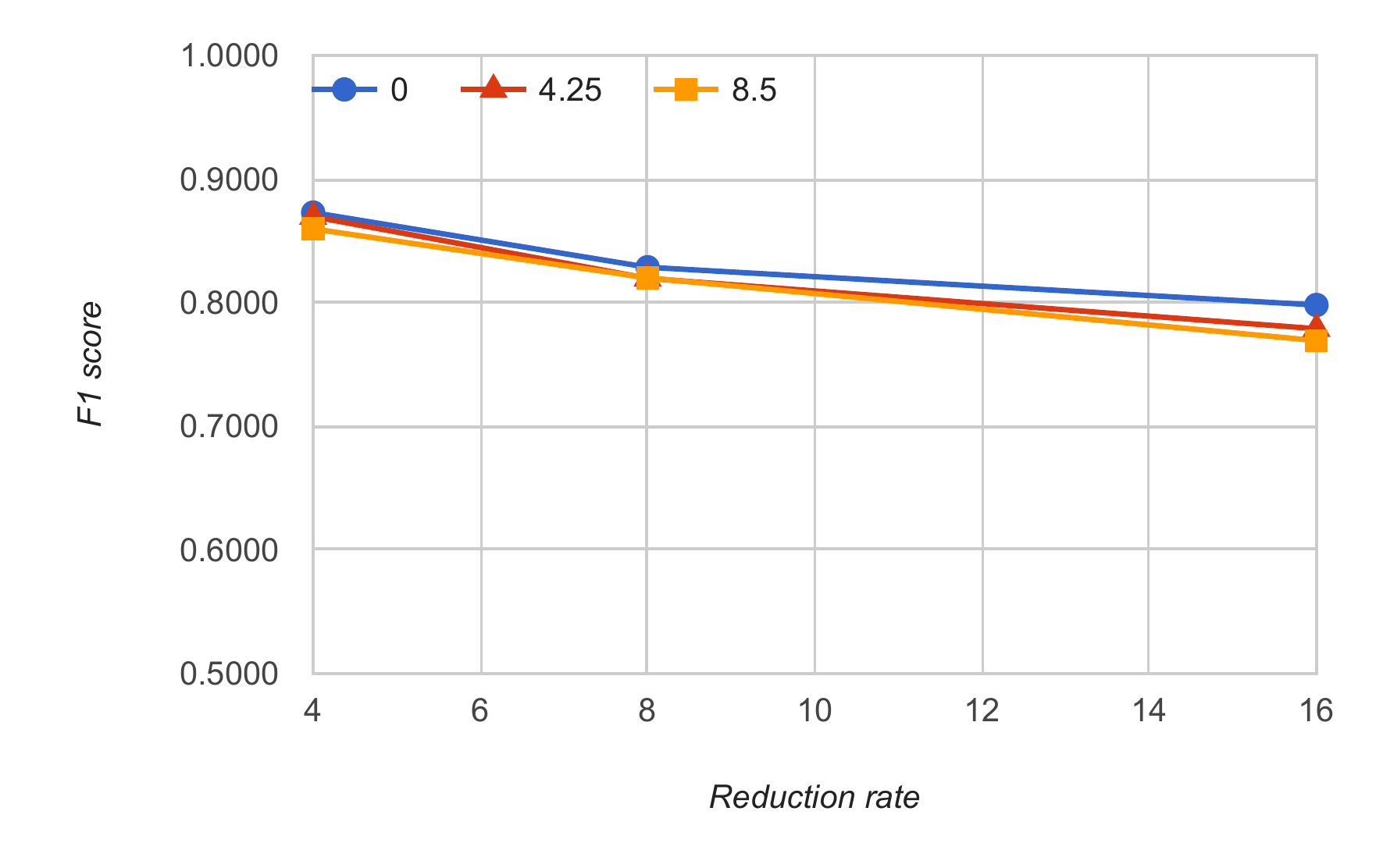}
    \vspace{-8pt}
    \caption{SP\textsubscript{M} + SVM}
\end{subfigure}
\begin{subfigure}{0.48\textwidth}
    \hspace{-12pt}
    \includegraphics[width=\textwidth]{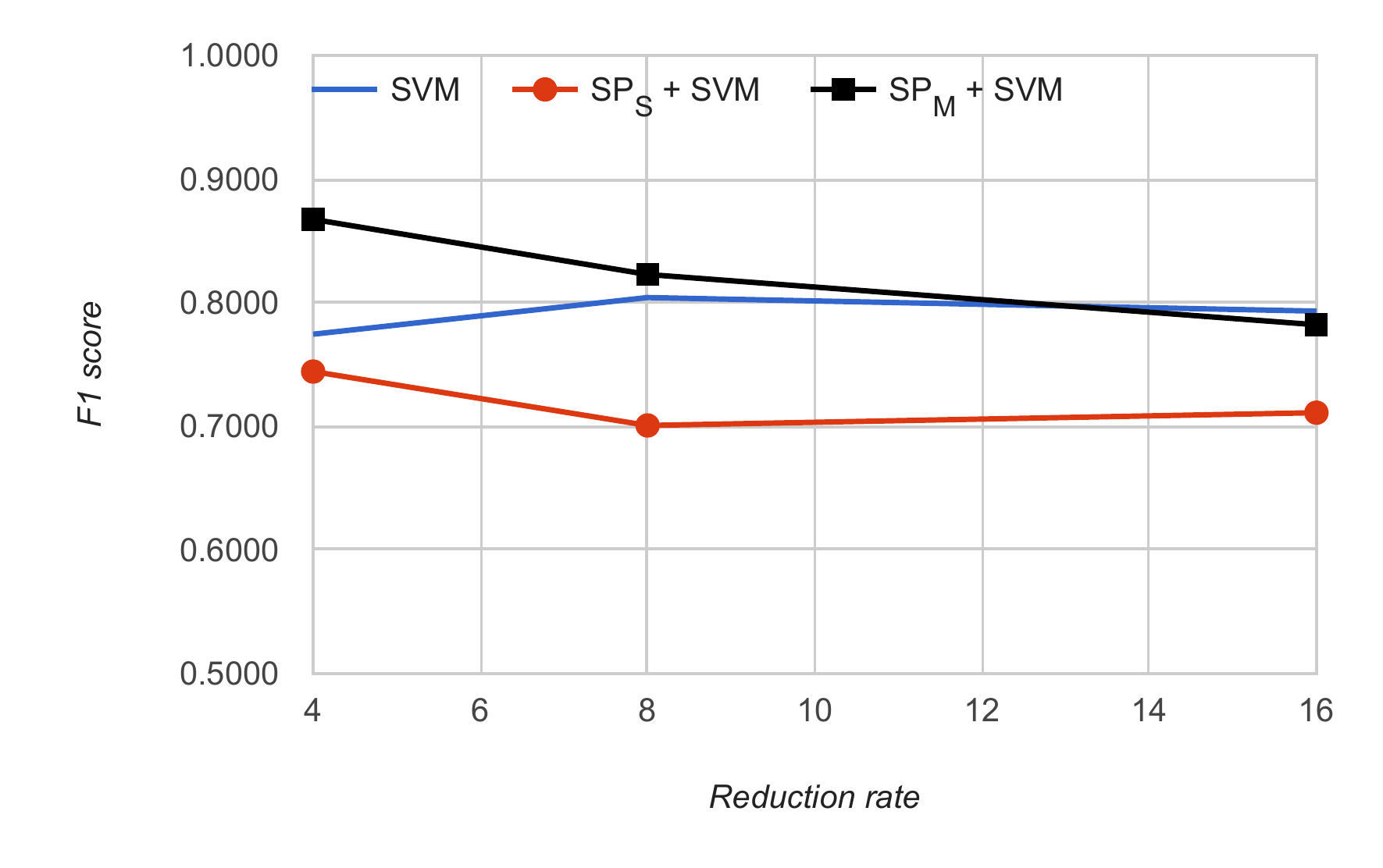}
    \vspace{-8pt}
    \caption{Average}
\end{subfigure}
\caption{F1 score as a function of different reduction rate ratio.}\label{fig:f1_reduction}
\end{figure}

Fig. \ref{fig:f1_reduction} shows superiority of the SP\textsubscript{M} + SVM setup for videos of lower reduction rate (i.e. containing more pixels). This results from the SP property of generating invariant representation which becomes more apparent when frames containing more data are provided as an input stream. It is worth emphasizing that increasing a number of pixels hinders performance of the pure SVM setup. It may be expected that further growth of input data size, along with application of a dedicated SP encoder \cite{Purdy}, may increase the superiority of the SP + SVM setup.

Fig. \ref{fig:f1_params} presents SP performance variation as a response to key parameter changes. The impact of the number of synapses and columns is the most significant, and it may be noticed that a rise of these parameters leads to better performance of the SP-based system. It turns out that value of $min\_overlap$ contributes significantly to the performance, as opposed to $winners\_set\_size$ which has very little or no impact on F1 score of the system.

As it is depicted in Fig. \ref{fig:f1_noise} -- \ref{fig:f1_params}, two different variants of SP introduction were examined, namely SP\textsubscript{S} and SP\textsubscript{M}. It is worth noting that 'Multiple HTMs' (SP\textsubscript{M}) tend to outperform the 'Single HTM' in terms of both F1 score and a pace of plateau convergence. However, this is achieved at the expense of a larger computational cost required for computing multiple instances of SP in 'Multiple HTMs' (SP\textsubscript{M}). 

According to the authors' knowledge it is hard to find papers which directly correspond to the research conducted in this work (i.e. video classification in noisy video streams). Nevertheless, we examined the following papers : \cite{Yue, KarpathyCVPR14, Zha, simonyan2014two-stream} which present results of video classification using a UCF-101 dataset. The best systems presented in those papers are based on various architectures of Convolutional Neural Networks (CNNs) and achieve accuracy of 80\% or more. It is worth emphasizing that despite similar performance in terms of the result quality, authors' test setup differs mostly in dataset used. Preliminary comparative experiments have been conducted with Temporal Stream ConvNet \cite{simonyan2014two-stream}. For UCF-11 dataset our solution achieves F1 score 0.6194, while Temporal Stream ConvNet (trained on the same amount of data) reached 0.1982; for shapes dataset, the best F1 score was 0.9544 while Temporal Stream ConvNet yielded results of 0.2908. It is worth emphasizing that both of the datasets used for the comparison experiments were relatively small. This amplified a skill of authors' solution to perform well despite being trained on small datasets. At the present development stage of the system, due to the computational requirements it would be hard for the authors to conduct tests with large datasets such as UCF-101.

\section{Conclusions and future work}
\label{section:conclusions}
This paper presents the experimental results of using an HTM--based system for object classification in noisy video streams. The authors showed that using SP in the video processing flow improves the object classification ratio by more than 10\% and achieves approximately 12 times the noise reduction for a video signal with  13\% distorted bits. It was determined that a rise of column and synapse number of SP has a substantial impact on the performance of the system and the best results were obtained for 256 and 4096 synapses and columns, respectively.

In future research, the authors plan to expand the work to test the system with more advanced benchmarks featuring real-live video streams. Such experiments were not done for this paper because we needed to verify operation of the video classification module in ideal conditions and adjust the basic parameters of the system. The authors are also going to replace the binarization operation with a custom--designed encoder and enhance the SVM classifier with a dedicated decoder. Experiments with several stacked SP layers extended with TP are also scheduled for future improvements of the system. However, in order to be able to examine performance of such configurations, the remaining computationally--exhaustive routines will have to be ported to OpenCL and the system will have to be deployed on platforms equipped with GPU or FPGA. This will enable conduction of experiments with video of a lower image reduction ratio.

\section*{Acknowledgment}
I would like to thank my wife Urszula Wielgosz for her huge contribution to the preparation of the paper.

\bibliography{bibliography,authors_works}

\end{document}